\documentclass[10pt,twocolumn,letterpaper]{article}

\usepackage[pagenumbers]{cvpr} 

\definecolor{cvprblue}{rgb}{0.21,0.49,0.74}
\usepackage[pagebackref,breaklinks,colorlinks,allcolors=cvprblue]{hyperref}
\usepackage{graphicx}
\usepackage{amsmath}
\usepackage{amssymb}
\usepackage{booktabs}
\usepackage{url}
\usepackage{algorithm}
\usepackage{algorithmic}
\usepackage[table]{xcolor} 
\usepackage{multirow}
\usepackage{array}
\usepackage{subcaption}
\usepackage{amsfonts}
\usepackage{enumitem}
\usepackage{framed}
\usepackage[accsupp]{axessibility}  

\def\paperID{42198	} 
\def\confName{CVPR}
\def\confYear{2026}

\title{RankOOD - Class Ranking-based Out-of-Distribution Detection}

\author{
    Dishanika Denipitiyage$^1$ \quad 
    Naveen Karunanayake$^1$ \quad
    Suranga Seneviratne$^1$ \quad
    Sanjay Chawla$^2$ \\
    {\tt\small \{dishanika.denipitiyage, naveen.karunanayake, suranga.seneviratne\}@sydney.edu.au}\\
    {\tt\small schawla@hbku.edu.qa}\\
    $^1$ The University of Sydney \quad 
    $^2$ Qatar Computing Research Institute, HBKU 
}

\begin{document}
\maketitle
\begin{abstract}
We propose  RankOOD, a rank-based Out-of-Distribution (OOD) detection approach based on training a model with the Placket-Luce loss, which is now extensively used for preference alignment tasks in foundational models. 
Our approach is based on the insight that with a deep learning model trained using the Cross Entropy Loss,  in-distribution (ID) class prediction induces a ranking pattern for each ID class prediction. The  RankOOD framework formalizes the insight by first extracting a rank list for each class using an initial classifier and
then uses another round of training with the Plackett-Luce loss, where the 
class rank, a fixed permutation for each class, is the predicted variable. An OOD example may get assigned with high
probability to an ID example, but the probability of it respecting the ranking
classification is likely to be small.  RankOOD achieves  SOTA performance on the near-ODD TinyImageNet evaluation benchmark, reducing FPR95 by $4.3\%$.

\end{abstract}

\section{Introduction}
\label{Sec:introduction_chap4}
\begin{figure}[t]
    \centering
    \includegraphics[width=0.9\linewidth]{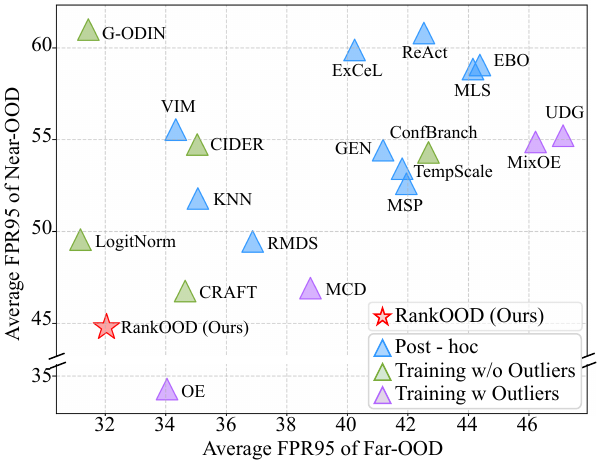}
    \caption{ The performance comparison of average FPR95 on Far-OOD (x-axis) and Near-OOD (y-axis). \textit{RankOOD outperforms all post-hoc and training methods without outliers. Compared to training methods with outliers, its performance is only second to the Outlier Exposure (OE) method.}}   
    \label{fig:fpr_near_vs_far}
    \vspace{-4mm}
\end{figure}
Despite their tremendous success, deployment of Deep Neural Networks (DNNs) in critical applications remains restricted due to the out-of-distribution (OOD) problem. For example, a vision classifier installed in an autonomous driving vehicle  may make an inaccurate but overconfident prediction for an object class not seen in the
training data. Accurate OOD detection continues to remain a fundamental unresolved problem not only in computer vision, but also in all machine learning~\cite{msp, yuan2023revisiting, shoeb2025out}.



Existing OOD detection approaches can be broadly divided into two categories: post-hoc methods and training-based methods. Post-hoc methods~\cite{msp, tempscale, ebo, react, vim} operate on pretrained models and extract OOD-related signals from their outputs or internal representations, offering simplicity and compatibility with existing networks. In contrast, training-based methods modify the learning process to improve the separability between in-distribution (ID) and OOD data. Among these, outlier-exposure techniques~\cite{oe, mcd, udg, mixoe} explicitly use auxiliary outlier datasets during training, while training without outliers methods enhance OOD robustness implicitly through regularization or objective design. Although post-hoc approaches are lightweight and maintain ID performance, training-based approaches typically achieve stronger OOD detection, often at the cost of reduced classification accuracy on ID samples.

Recently, a new line of work is emerging on addressing the OOD problem, which
is post-hoc and includes elements of post-training. The key insight is the following:
\vspace{-1mm}
\FrameSep4pt
\begin{framed}A DNN classifier trained for single-class prediction often naturally induces
a rank order across classes. An OOD example may be overconfidently assigned to
a class label but is unlikely to respect the associated rank order.  Thus, if we
can train a classifier to strengthen the rank order, then we can achieve
both accurate in-distribution prediction and OOD detection.
\end{framed}
\vspace{-1mm}

Several works leverage class ranking patterns as a discriminative signal for OOD detection~\cite{openmax,karunanayake2025craft, excel}. For instance, CRAFT~\cite{karunanayake2025craft} fine-tunes pre-trained models to learn class-specific ranking distributions among prediction scores. CRAFT demonstrates that ID samples exhibit more deterministic inter-class ranking relationships, while OOD samples disrupt these patterns. By modeling these relationships via class-dependent probability mass functions (PMFs) and measuring their divergence during inference, CRAFT achieves strong OOD performance. However, this fine-tuning process introduces architectural modifications and ignores the relative ordering of the class ranks. 

In this work, we propose RankOOD, a rank-based, listwise OOD detection framework that eliminates the need for fine-tuning or a separate network while preserving the discriminative power of ranking structures. 
Unlike CRAFT, which models each class ranking as a 2D C×C PMF matrix (where 
C is the number of classes), RankOOD directly operates on the raw logits of a pre-trained network and orders them according to predefined rank labels. We employ a Listwise Maximum Likelihood Estimation (ListMLE)~\cite{xia2008listwise} objective to learn the rank structure. Unlike pointwise or pairwise objectives, ListMLE treats the entire output list jointly and assigns a probability to each permutation via the Plackett–Luce model~\cite{marden1996analyzing}, maximizing the likelihood of the observed ranking. This allows the model to capture relative ordering dependencies among classes, reflecting their competitive nature in the final decision. Conceptually, our method captures global listwise consistency, modeling how class scores interact within a shared ranking space. This perspective not only simplifies the training pipeline, since no additional fine-tuning or outlier exposure is required, but also provides a canonical ranking structure for OOD detection. We summarize the following contributions,
\begin{itemize}
    \item We propose RankOOD, a novel method that detects OOD inputs by learning and analyzing class-wise rank order patterns directly from model outputs without fine-tuning or auxiliary outliers.
    \item We introduce the novel use of a ListMLE objective derived from the Plackett–Luce (PL) model~\cite{marden1996analyzing}, to model entire output rankings for OOD detection, effectively capturing vital inter-class dependencies ignored by prior methods.
    \item We validate RankOOD by comparing $34$ existing methods conducted in the OpenOOD environment~\cite{openood}. Our method consistently ranks within the top two in near-OOD detection and the top three for far-OOD setting, demonstrating the strong consistency among the evaluated methods. Notably, on the TinyImageNet near-OOD, RankOOD achieves SOTA performance, reducing FPR95 by 4.3\% relative to the strongest baseline.
\end{itemize}

\section{Related Work}
\label{Sec:related_work_chap4
}

OOD detection methods for image classifiers are primarily grouped into two categories: i) post-hoc inference methods~\cite{msp,mds,ebo} and ii) training-based methods~\cite{logitnorm,rotpred,oe}.  

\noindent\textbf{Post-hoc inference methods:} These methods utilize a standard classifier (usually trained with cross-entropy) and compute an OOD score from its outputs or intermediate features. A simple baseline is the Maximum Softmax Probability (MSP) detector, which uses the largest softmax score as the anomaly score~\cite{msp}. Other classic scores include the Mahalanobis distance~\cite{mds}, which measures deviations from class-conditional feature statistics, and the energy score~\cite{ebo}, defined as the negative log-sum-exp of the logits. More recent methods aim to further improve the separability between ID and OOD samples by refining DNN outputs or activations through techniques such as rectification~\cite{react}, shaping~\cite{ash}, and sparsification~\cite{dice}.

\noindent\textbf{Training-based methods:} Following the OpenOOD taxonomy~\cite{openood}, training-based methods can be further grouped into methods without outliers~\cite{logitnorm,godin,cider} and with outliers~\cite{oe,mixoe,mcd}. \textit{Training methods without outliers} modify the learning objective using only ID data to improve calibration and robustness. RotPred~\cite{rotpred} introduces an auxiliary rotation prediction task, leveraging self-supervision to encourage feature diversity that aids OOD detection. CSI~\cite{csi} applies contrastive self-supervised learning to cluster ID samples tightly in embedding space. LogitNorm~\cite{logitnorm} enforces a fixed norm on the output logits during training, preventing overconfident predictions and substantially improving post-hoc OOD separability. These methods offer strong performance without relying on external data, making them practical for closed-set training scenarios.

Conversely, when auxiliary outlier data are available, models can explicitly learn to assign low confidence to them (i.e., \textit{training methods with outliers}). The outlier exposure (OE) method~\cite{oe} introduces a loss term that penalizes overconfident predictions on a diverse auxiliary dataset, encouraging uniform output distributions for OOD samples. Building on OE, MixOE~\cite{mixoe} interpolates ID and OOD samples to generate mixed examples that regularize model confidence. Other variants, such as MCD~\cite{mcd} and UDG~\cite{udg}, leverage ensemble disagreement or unsupervised clustering to synthesize pseudo-OOD data. 

Each category of methods has its own strengths and trade-offs. Post-hoc approaches are simple to apply and require no retraining, but their effectiveness depends heavily on the quality and calibration of the underlying model. Training-based methods without outliers tend to be more robust and data-efficient, though they may struggle when confronted with unfamiliar, real-world anomalies. Outlier-assisted methods often achieve the strongest results but risk overfitting to the seen outlier data and rely on access to sufficiently diverse and representative outlier examples, which may not always be practical.

\noindent\textbf{Class rank-based OOD detection} A recent line of research has leveraged class-rank information to improve OOD detection. For example, ExCeL~\cite{excel} computes a post-hoc OOD score by combining the maximum logit with a \textit{class rank signature} that captures the probability of each class appearing in subsequent ranks. Extending this idea, CRAFT~\cite{karunanayake2025craft} enhances OOD detection by fine-tuning a pre-trained classifier to sharpen its learned class-rank patterns, which are modeled as PMFs. \textit{Our work, RankOOD further advances this direction by optimizing class-wise rank structures under the Plackett–Luce formulation~\cite{marden1996analyzing}, using standard vision backbones without architectural changes.}

\section{Methodology}
\label{Sec:methodology_chap4}


In this section, we present the overall methodology of RankOOD, which consists of three main steps: i) using a pre-trained model on ID data, we derive a canonical class ranking for each class by solving a rank assignment problem (Sec.~\ref{sec:ilp}); ii) we then train a new classifier on the ID data using the canonical class rankings as ground truth and the ListMLE loss, referred to as RankOOD-T, the training process of our framework (Sec.~\ref{sec:rankood_t}); and iii) during inference, we compute an OOD score (termed RankOOD-S) based on the predicted ranks and their deviation from the canonical rankings, under the premise that OOD samples deviate more from the canonical class ranking patterns (Sec.~\ref{sec:rankood_s}).

\subsection{Finding Canonical Class Ranks}
\label{sec:ilp}

Using a pre-trained model on in-distribution train data, we first compute a \textit{Rank Probability Matrix} (RPM)~\cite{karunanayake2025craft} for each class $c \in C$, which comprises the probability mass functions (PMFs) across all rank positions, estimated from ID samples that are correctly predicted as class $c$.
Specifically, for each class $c$, an element $p^c_{i,j}$ within the RPM, 
$P^c \in \mathbb{R}^{C\times K}$ denotes the probability that class $i$ appears at the $j^{th} \in K$ rank when the input is classified as class 
$c$. Thus, each column of the matrix $P^c_j$ represents a PMF over ID classes corresponding to a particular rank position $j$. 

To obtain a consistent canonical class ranking for each top-rank class, we solve an Integer Linear Programming (ILP) problem. That is, we formulate a 0-1 integer linear program that selects one class per rank, ensuring uniqueness and maximizing the total assignment probability. For each class, this yields a consistent ranking that best reflects the trained preference structure.

Given the class $c$, and its rank probability matrix, $P^c \in \mathbb{R}^{C\times K}$, we introduce binary decision variables,
\vspace{-0.205mm}
\[    x_{i,j}^c = 
\begin{cases}
1, & \text{if class } i \text{ is assigned to rank } j,\\
0, & \text{otherwise.}
\end{cases}
\]
We compute the consensus ranking by solving the following 0-1 integer linear program:
\begin{equation}
    \max_{x} \sum_{i=1}^{C} \sum_{j=1}^{K} x_{i,j}^c p_{i,j}
    \vspace{-4mm}
\end{equation}
subject to
\[
\vspace{-1mm}
\sum_{i=1}^{C} x_{i,j}^c = 1 \space\space \forall j\in [1, K], \quad
\sum_{j=1}^{K} x_{i,j}^c \le 1 \space\space \forall i\in [1, C]
\]
This yields a valid ranking permutation that maximizes the joint probability of the $K$ ranking structure under the model. Note that the first constraint ensures only one class is selected per rank, and the second constraint ensures that a class is selected at most once, across all considered ranks.\\  

\noindent{{\bfseries Example:}} Consider this worked example in a four-class classification problem as shown in Table~\ref{Tab:RPMExample}. We consider the 100 correctly classified samples of Class 2. 

Since we are considering only 100 correctly classifying samples of Class 2, the frequency $f^2_{ij}$ for (2,0), i.e., the number of samples where Class 2 is in the $0^{th}$ rank is 100. This also means, there are no samples with any other class in rank 0, i.e., $f^2_{i0}\; \forall \; i\backslash \{2\}$ is 0. Also means that Class 2 can not appear in any other position than rank-0, i.e., $f^2_{2j}\; \forall \; j \backslash \{0\}$ is 0. 

Consider the other $f^2_{ij}$ values as filled in the table. Note that the $j^{th}$ column sum, i.e., $\sum_{j=1}^{4}f^2_{ij}$ is 100. Next, the relative frequency can be converted to probabilities, $p^2_{ij}$, as shown inside brackets in the table. The shaded cells of this table, ignoring the trivial row and column for class 2, form the RPM $\in 4 \times 4$ for Class 2.

Note that in this example, for Class 2, rank 1 is dominated by Class 1, and rank 2 is dominated by Class 3, and rank 3 is dominated by Class 4. Though the canonical class is clear cut in this example, in practice, as the number of classes increases, RPMs get noisier, leading to ties between classes in some ranks. Our ILP solution ensures that the best representative ranking order is selected for each class. 

\begin{table}[h]
\footnotesize
\centering
\renewcommand{\arraystretch}{1.2}
\setlength{\tabcolsep}{10pt}
\begin{tabular}{c c|c|c|c|c|c}
\multicolumn{2}{c}{} & \multicolumn{5}{c}{\textbf{Rank (j)}} \\[2pt]
\multicolumn{2}{c}{} & 0& 1 & 2 & 3 &  \\ 
\cline{3-6} 
\multirow{5}{*}{\rotatebox{90}{\textbf{Class (i)}}} 
 & 1 & 0 & \cellcolor{lightgray} 80 (0.80) &  \cellcolor{lightgray} 15 (0.15) &  \cellcolor{lightgray} 5 (0.05)  \\ 
\cline{2-6} 
 & 2 & 100 & 0 & 0 & 0  \\ 
\cline{2-6}
 & 3 &  0&  \cellcolor{lightgray}10 (0.10) &  \cellcolor{lightgray} 75 (0.75) &  \cellcolor{lightgray} 15 (0.15)  \\ 
\cline{2-6}
 & 4 &  0&  \cellcolor{lightgray} 10 (0.10) &  \cellcolor{lightgray} 10 (0.10) &  \cellcolor{lightgray} 80 (0.80)  \\ 
\cline{2-6}
\end{tabular}
\caption{An example RPM for Class 2, in a four class classification problem}
\label{Tab:RPMExample}
\vspace{-3mm}
\end{table}

\subsection{Ordered Preference Learning}
\label{sec:rankood_t}

Next, leveraging the ranking information generated through ILP, we train a model using the hybrid objective combining cross‐entropy (CE) loss and Listwise Maximum Likelihood Estimation (ListMLE)~\cite{xia2008listwise}. While CE encourages the model to predict the correct top class, ListMLE enforces consistency across the entire ordered label sequence by maximizing the likelihood of the observed ranking under the Plackett–Luce formulation~\cite{marden1996analyzing}. The ListMLE loss is defined as: 
\begin{equation}
    \mathcal{L}_{\text{ListMLE}} = -\sum_{i=0}^{K-1}( l_{\pi_i} - log( \sum_{j=i}^{K-1} \exp(l_{\pi_j})))
    \label{eq:listmle}
\end{equation}
equivalently expressed under the Plackett–Luce model as
\begin{equation}
    \mathcal{P}(\pi|l) = \prod_{i=0}^{K-1} \frac{\exp(l_{\pi_i})}{\sum_{j=i}^{K-1} \exp(l_{\pi_j})}, \quad \mathcal{L}_{\text{ListMLE}} = -log [\mathcal{P}(\pi|l)]
    \label{eq:pl_prob}
\end{equation}

Here, $\pi=(\pi_0,\pi_1,\ldots,\pi_{K-1})$ denote the ground-truth ranking where $\pi_i$ is the class index assigned to rank $i$ and $l_{\pi_i}$ denote the logits of the class at rank $i$ produced by the model.
The Eq.~\ref{eq:pl_prob} defines a parameterized exponential probability distribution over all the permutations given the predicted result by the model $\mathcal{F}_\theta$, and defines the loss function (Eq.~\ref{eq:listmle}) as the negative log likelihood of the ground truth ranks. This formulation enforces relative ordering ($l_{\pi_0}>\ldots>l_{\pi_i}> \ldots >l_{\pi_{K-1}}$) consistency across the entire ranked list rather than optimizing only the top-1 classification decision. Unlike conventional CE loss, which encourages only the correct label to have maximal score, ListMLE optimizes the full permutation likelihood, ensuring a coherent and structured logit hierarchy. This property makes ListMLE particularly suitable when the model must preserve rich class-relation structure and when OOD detection depends on stable, interpretable score ordering beyond the top prediction. 
We empirically show that it is not necessary to train the model on the entire rank sequence when only a subset of ranks is sufficient for the downstream OOD score and we selected top-$k$ ranks and lowest-$k$ ranks. However, training on a subset of ranks alone does not constrain the absolute ordering of unsupervised middle positions, nor does it guarantee that the $l_{\pi_0}$
is globally maximal among all $C$ logits. Therefore, to ensure that the predicted top class is indeed the argmax of the full logit vector, we therefore complement the ListMLE loss with a cross-entropy term, 
\begin{equation}
    \label{eq: loss}
    \mathcal{L}_{RankOOD-T} = \mathcal{L}_{CE} + \alpha\mathcal{L}_{\text{ListMLE}}
\end{equation}
The hyperparameter $\alpha$ balances the trade-off between the two objectives. 

\subsection{Detecting OOD Samples}
\label{sec:rankood_s}
To effectively distinguish ID samples from OOD inputs, we leverage the structure of rank-based logits learned by our RankOOD-T model. First, we derive class-dependent logit threshold profiles from confident training samples, capturing characteristic logit decay across ranks for correctly classified samples. Second, at test time, we compute a RankOOD-S that penalizes deviations from both the expected ranking order and the reference logit thresholds.\\
\noindent\textbf{Logit threshold Profile - $Ref$:}
We construct a reference threshold profile for each class based on the logit values of correctly predicted training samples at rank-$0$. Among these, we further retain only those samples that satisfy the condition that at least $N$ rank positions yield correct predictions, where $N$ is chosen such that each class has at least one qualifying training sample. For each rank position $i$, we compute the class-specific reference logit threshold, ${Ref}^{c}_{i}$ as the empirical $95^{th}$-percentile of logits. In Sec.~\ref{sec:ablation} we explain the percentile selection using the validation set provided by the OpenOOD benchmark~\cite{openood}.\\
\noindent\textbf{RankOOD-S: }
Let $x$ denote the logit vector for a given test input,
we first determine its predicted class label $\hat{c}$ (i.e., the class related to the highest logit value in $x$) and obtain the expected ranking order $\pi^{\hat{c}}$ for the predicted class $\hat{c}$. Let $\Bar{\pi}$ denote the ranking predicted by the input sample. For each rank position $i$, we introduce a penalty term if the model misorders the rank, $\Bar{\pi}_{i}\neq \pi^{\hat{c}}_{i}$. Since the Plackett–Luce objective couples rank $i$ with all preceding ranks $[0, i]$, an incorrect rank at position $i$ contributes to a deviation in logits in previous ranks. Therefore, we define a cumulative margin penalty on logits in each expected rank $i$:
\begin{equation}
    \label{eq:penalty}
    \delta_{\pi^{\hat{c}}_i} = \gamma^r, \quad r={\sum_{j=i}^{K-1}1[\pi^{\hat{c}}_{j} \neq \Bar{\pi}_{j}]} 
\end{equation}
where $\gamma \geq 1$. Finally, the RankOOD-S measures the weighted deviation of penalized logits from the class-specific reference threshold profile:
\begin{align}
    \label{eq:ood_score}
    \text{RankOOD-S} = \sum_{i=0}^{K-1} w_i \log \left(\operatorname{softmax}(\mathbf{u})\right)_i \\
\text{where} \quad u_i=\frac{x_{{\pi^{\hat{c}}_i}}}{\delta_{{\pi^{\hat{c}}_i}}} - Ref_i^{\hat{c}} \quad  \forall i\in\{0,\ldots,K-1\} \notag
\end{align}

where $w_i$ is learned via linear regression on validation data to maximize separation between ID and OOD samples. Intuitively, ID samples are expected to exhibit a high rank-0 logit value followed by a monotonically decreasing logit sequence that maintains the logit ordering structure across ranks.
Since ListMLE in RankOOD-T loss enforces a sequential scoring dependency, a single mid-rank violation reveals inconsistency in the entire confidence trajectory, making the method inherently sensitive to OOD distortions. An example of RankOOD-S is provided in the Appendix~A.5.


\section{Experiments}
\label{_chap4}


\textbf{Datasets: }We use CIFAR-10/100~\cite{cifar100} and ImageNet-200 (a.k.a., TinyImageNet)~\cite{tin} as ID datasets for our experiments. Following the OpenOOD benchmark~\cite{openood}, we evaluate each ID dataset against both near-OOD and far-OOD test datasets.
Specifically, for CIFAR-10, the near-OOD datasets include CIFAR-100 and TinyImageNet, while for CIFAR-100, CIFAR-10 and TinyImageNet are used as near-OOD. For both CIFAR-10 and CIFAR-100, the far-OOD group comprises MNIST~\cite{mnist}, SVHN~\cite{svhn}, Textures~\cite{textures}, and Places365~\cite{places}. For ImageNet-200, SSB-hard~\cite{ssb_hard} and NINCO~\cite{ninco} serve as near-OOD datasets, whereas iNaturalist~\cite{inaturalist}, Textures~\cite{textures}, and OpenImage-O~\cite{vim} form the far-OOD group. To ensure consistency, we adopt the same training, validation, and testing splits provided by the OpenOOD benchmark.

\noindent\textbf{Evaluation Metrics: }Similar to previous work, we use two standard metrics to evaluate OOD detection performance: (1) FPR95, the false positive rate for OOD samples when the true positive rate for ID samples is 95\%; and (2) AUROC, the area under the receiver operating characteristic curve, which measures the detector’s ability to distinguish between ID and OOD samples across all thresholds. We report the mean and standard deviation of these metrics over three independent runs with different random seeds.

\noindent\textbf{Baselines: }We compare RankOOD against a comprehensive set of $34$ OOD detection methods. We categorize the baselines into three major categories based on the inference methods and the training methods: post-hoc methods, training methods without outliers, and training methods with outliers. Among post-hoc methods, we include classical approaches such as MSP~\cite{msp} and Energy-based OOD detection (EBO)~\cite{ebo}, as well as more recent techniques like ReAct~\cite{react}, ASH~\cite{ash}, and GEN~\cite{gen}. Training-based methods without auxiliary outlier data include ConfBranch~\cite{confbranch}, G-ODIN~\cite{godin}, and LogitNorm~\cite{logitnorm}. Conversely, methods that utilize auxiliary outlier data for training include Outlier Exposure (OE)~\cite{oe}, MCD~\cite{mcd}, UDG~\cite{udg}, and MixOE~\cite{mixoe}. Finally, we compare our method against two existing rank-based methods, ExCeL~\cite{excel} and CRAFT~\cite{karunanayake2025craft}. 

\noindent\textbf{Implementation Details: }For all experiments, we use pre-trained ResNet-18 models~\cite{resnet18} as the backbone. For both CIFAR datasets, each model is trained for 500 epochs, while the TinyImageNet model is trained for 300 epochs using RankOOD-T loss. All models use the SGD optimizer with a momentum of 0.9, an initial learning rate of 0.1, and a cosine annealing decay schedule. In our method, we perform hyperparameter sweeps over the loss weighting parameter $\alpha$ in Eq.~\ref{eq: loss} as well as over the reference logit threshold, $Ref$. We set $\alpha$ to $0.8, 1.0,$ and $0.5$ for CIFAR-10, CIFAR-100, and TinyImageNet, respectively, and select the $95^{th}$ percentile of the logit distribution as the reference logit threshold for all datasets. The CIFAR-10 network is trained using all 10 ranks, whereas the CIFAR-100 and TinyImageNet models are trained using the top 10 and bottom 10 ranks generated through the ILP procedure. Hyperparameter search details are provided in Sec.~\ref{sec:ablation}. Additional details on GPU hours, ILP complexity, and ILP runtime are provided in the Appendix~A.4.

\section{Results}
\label{sec:results_chap4}

We summarize near-OOD results in Tab.~\ref{tab:nearood} and far-OOD results in Tab.~\ref{tab:farood} on CIFAR-10/100 and TinyImageNet. For brevity, we present average near- and far-OOD performance across all benchmarks for each ID dataset (Sec.~\ref{sec:sota_results}), as well as the overall OOD detection performance. Complete results for all $34$ methods, including per-dataset performance, are provided in the Appendix~A.1 and A.2.

\subsection{Comparison with SOTA Methods}
\label{sec:sota_results} 
Across the average performance over all benchmarks, RankOOD consistently ranks among the top three methods, achieving the second-best near-OOD ({\bf cf.}~Tab.~\ref{tab:nearood}) in terms of both AUROC and FPR95. RankOOD is also the third-best far-OOD  ({\bf cf.}~Tab.~\ref{tab:farood}) performance. Similarly, RankOOD outperforms all prior rank-based approaches, CRAFT~\cite{karunanayake2025craft} and ExCel~\cite{excel} in both near- and far-OOD settings, achieving an average FPR95 reduction of $7.51$\% for far-OOD and $4.21\%$ for near-OOD, while outperforming both CIFAR-100 and TinyImageNet datasets and lies on-par with CIFAR-10 performance. 

In the near-OOD setting~({\bf cf.} Tab.~\ref{tab:nearood}), the only method surpassing RankOOD is OE~\cite{oe}, which benefits from access to outlier samples during training—an assumption RankOOD does not make. Notably, in the near-OOD setting, RankOOD achieves state-of-the-art performance on TinyImageNet, improving AUROC by 0.50\% and reducing FPR95 by 4.3\% compared to the strongest baseline. This highlights the value of incorporating semantic rank structure when detecting OOD samples in high-cardinality label spaces. Furthermore, among methods that do not utilize outliers, RankOOD obtains the best FPR95 on CIFAR-100, outperforming GEN~\cite{gen} by 3.36\%. 

As shown in Tab.~\ref{tab:farood}, despite being on par with top five methods for CIFAR-10, RankOOD ranks second on CIFAR-100 with 83.63\% AUROC and 47.44\% FPR95, trailing only G-ODIN~\cite{godin}. Nonetheless, RankOOD outperforms G-ODIN by 8.12\% in TinyImageNet FPR95, and G-ODIN performs worst in near-OOD setting for all the datasets. Moreover, in far-OOD setting, RankOOD outperforms all outlier-exposure-based methods on CIFAR-100 and TinyImageNet across all metrics.

\begin{table*}[t]
\scriptsize
\centering
\caption{\centering Performance comparison in \textit{near-OOD detection}. For each column, the top five methods are marked in \textbf{bold}. }
\label{tab:nearood}
\begin{tabular}{lcc|cc|cc|cc}
\hline
Method & \multicolumn{2}{c|}{CIFAR-10} & \multicolumn{2}{c|}{CIFAR-100} & \multicolumn{2}{c|}{ImageNet-200} & \multicolumn{2}{c}{Average} \\
 & AUROC $\uparrow$ & FPR95 $\downarrow$ & AUROC $\uparrow$ & FPR95 $\downarrow$ & AUROC $\uparrow$ & FPR95 $\downarrow$ & AUROC $\uparrow$ & FPR95 $\downarrow$ \\
\hline
\multicolumn{9}{c}{Post-hoc inference methods} \\ 
\hline
MSP~\cite{msp} & 88.03 ± 0.25 & 48.17 ± 3.92 & 80.27 ± 0.11 & \textbf{54.80 ± 0.33} & 83.34 ± 0.06 & 54.82 ± 0.35 & 83.88 & 52.60 \\
TempScale~\cite{tempscale} & 88.09 ± 0.31 & 50.96 ± 4.32 & 80.90 ± 0.07 & \textbf{54.49 ± 0.48} & \textbf{83.69 ± 0.04} & 54.82 ± 0.23 & 84.23 & 53.42 \\
RMDS~\cite{rmds} & 89.80 ± 0.28 & 38.89 ± 2.39 & 80.15 ± 0.11 & 55.46 ± 0.41 & 82.57 ± 0.25 & \textbf{54.02 ± 0.58} & 84.17 & \textbf{49.46} \\
EBO~\cite{ebo} & 87.58 ± 0.46 & 61.34 ± 4.63 & \textbf{80.91 ± 0.08} & 55.62 ± 0.61 & 82.50 ± 0.05 & 60.24 ± 0.57 & 83.66 & 59.07 \\
ReAct~\cite{react} & 87.11 ± 0.61 & 63.56 ± 7.33 & 80.77 ± 0.05 & 56.39 ± 0.34 & 81.87 ± 0.98 & 62.49 ± 2.19 & 83.25 & 60.81 \\
MLS~\cite{mls} & 87.52 ± 0.47 & 61.32 ± 4.62 & \textbf{81.05 ± 0.07} & 55.47 ± 0.66 & 82.90 ± 0.04 & 59.76 ± 0.59 & 83.82 & 58.85 \\
VIM~\cite{vim} & 88.68 ± 0.28 & 44.84 ± 2.31 & 74.98 ± 0.13 & 62.63 ± 0.27 & 78.68 ± 0.24 & 59.19 ± 0.71 & 80.78 & 55.55 \\
KNN~\cite{knn} & 90.64 ± 0.20 & 34.01 ± 0.38 & 80.18 ± 0.15 & 61.22 ± 0.14 & 81.57 ± 0.17 & 60.18 ± 0.52 & 84.13 & 51.80 \\
ASH~\cite{ash} & 75.27 ± 1.04 & 86.78 ± 1.82 & 78.20 ± 0.15 & 65.71 ± 0.24 & 82.38 ± 0.19 & 64.89 ± 0.90 & 78.62 & 72.46 \\
GEN~\cite{gen} & 88.20 ± 0.30 & 53.67 ± 3.14 & \textbf{81.31 ± 0.08} & \textbf{54.42 ± 0.33} & \textbf{83.68 ± 0.06} & 55.20 ± 0.20 & \textbf{84.40} & 54.43 \\
ExCeL~\cite{excel} & 86.89 ± 0.23 & 66.55 ± 0.43 & 80.70 ± 0.06 & 55.21 ± 0.56 & 82.40 ± 0.04 & 57.90 ± 0.40 & 83.33 & 59.89 \\
\hline
\multicolumn{9}{c}{Training methods without outliers} \\
\hline
\rowcolor{gray!30}

RankOOD (ours) & 90.21 ± 0.41 & \textbf{31.72 ± 0.67} & 80.67 ± 0.40 & \textbf{52.59 ± 0.75} & \textbf{85.30 ± 0.18} & \textbf{50.05 ± 0.16} & \textbf{85.39} & \textbf{44.79} \\
\hline
CRAFT~\cite{karunanayake2025craft} & \textbf{91.11 ± 0.04} & \textbf{31.94 ± 1.41} & 80.90 ± 0.33 & 53.73 ± 0.62 & \textbf{83.65 ± 0.41} & \textbf{54.62 ± 0.57} & \textbf{85.22} & 46.76 \\
ConfBranch~\cite{confbranch} & 89.84 ± 0.24 & 31.28 ± 0.66 & 71.60 ± 0.62 & 70.21 ± 0.83 & 79.10 ± 0.24 & 61.44 ± 0.34 & 80.18 & 54.31 \\
G-ODIN~\cite{godin} & 89.12 ± 0.57 & 45.54 ± 2.52 & 77.15 ± 0.28 & 67.58 ± 0.98 & 77.28 ± 0.10 & 69.87 ± 0.46 & 81.18 & 61.00 \\
LogitNorm~\cite{logitnorm} & \textbf{92.33 ± 0.08} & \textbf{29.34 ± 0.81} & 78.47 ± 0.31 & 62.89 ± 0.57 & 82.66 ± 0.15 & 56.46 ± 0.37 & \textbf{84.49} & \textbf{49.56} \\
CIDER~\cite{cider} & \textbf{90.71 ± 0.16} & 32.11 ± 0.94 & 73.10 ± 0.39 & 72.02 ± 0.31 & 80.58 ± 1.75 & 60.10 ± 0.73 & 81.46 & 54.74 \\
\hline
\multicolumn{9}{c}{Training methods with outliers} \\
\hline
OE~\cite{oe} & \textbf{94.82 ± 0.21} & \textbf{19.84 ± 0.95} & \textbf{88.30 ± 0.10} & \textbf{30.73 ± 0.11} & \textbf{84.84 ± 0.16} & \textbf{52.30 ± 0.67} & \textbf{89.32} & \textbf{34.29} \\
MCD~\cite{mcd} & \textbf{91.03 ± 0.12} & \textbf{30.17 ± 0.06} & 77.07 ± 0.32 & 55.88 ± 0.85 & 83.62 ± 0.09 & \textbf{54.71 ± 0.83} & 83.91 & \textbf{46.92} \\
UDG~\cite{udg} & 89.91 ± 0.25 & 35.34 ± 0.95 & 78.02 ± 0.10 & 61.42 ± 0.48 & 74.30 ± 1.63 & 68.89 ± 1.72 & 80.74 & 55.22 \\
MixOE~\cite{mixoe} & 88.73 ± 0.82 & 51.45 ± 7.78 & \textbf{80.95 ± 0.20} & 55.22 ± 0.49 & 82.62 ± 0.03 & 57.97 ± 0.40 & 84.10 & 54.88 \\
\hline
\end{tabular}
\end{table*}


\begin{table*}[t!]
\scriptsize
\centering
\caption{\centering Performance comparison in \textit{far-OOD detection}. For each column, the top five methods are marked in \textbf{bold}.}
\label{tab:farood}
\begin{tabular}{lcc|cc|cc|cc}
\hline
Method & \multicolumn{2}{c|}{CIFAR-10} & \multicolumn{2}{c|}{CIFAR-100} & \multicolumn{2}{c|}{ImageNet-200} & \multicolumn{2}{c}{Average} \\
 & AUROC $\uparrow$ & FPR95 $\downarrow$ & AUROC $\uparrow$ & FPR95 $\downarrow$ & AUROC $\uparrow$ & FPR95 $\downarrow$ & AUROC $\uparrow$ & FPR95 $\downarrow$ \\
\hline
\multicolumn{9}{c}{Post-hoc inference methods} \\
\hline
MSP~\cite{msp} & 90.73 ± 0.43 & 31.72 ± 1.84 & 77.76 ± 0.44 & 58.70 ± 1.06 & 90.13 ± 0.09 & 35.43 ± 0.38 & 86.21 & 41.95 \\
TempScale~\cite{tempscale} & 90.97 ± 0.52 & 33.48 ± 2.39 & 78.74 ± 0.51 & 57.94 ± 1.14 & 90.82 ± 0.09 & 34.00 ± 0.37 & 86.84 & 41.81 \\
RMDS~\cite{rmds} & 92.20 ± 0.21 & 25.35 ± 0.73 & \textbf{82.92 ± 0.42} & 52.81 ± 0.63 & 88.06 ± 0.34 & 32.45 ± 0.79 & 87.73 & 36.87 \\
EBO~\cite{ebo} & 91.21 ± 0.92 & 41.69 ± 5.32 & 79.77 ± 0.61 & 56.59 ± 1.38 & 90.86 ± 0.21 & 34.86 ± 1.30 & 87.28 & 44.38 \\
ReAct~\cite{react} & 90.42 ± 1.41 & 44.90 ± 8.37 & 80.39 ± 0.49 & 54.20 ± 1.56 & \textbf{92.31 ± 0.56} & 28.50 ± 0.95 & 87.71 & 42.53 \\
MLS~\cite{mls} & 91.10 ± 0.89 & 41.68 ± 5.27 & 79.67 ± 0.57 & 56.73 ± 1.33 & 91.11 ± 0.19 & 34.03 ± 1.21 & 87.29 & 44.15 \\
VIM~\cite{vim} & 93.48 ± 0.24 & 25.05 ± 0.52 & 81.70 ± 0.62 & \textbf{50.74 ± 1.00} & 91.26 ± 0.19 & \textbf{27.20 ± 0.30} & 88.81 & \textbf{34.33} \\
KNN~\cite{knn} & 92.96 ± 0.14 & 24.27 ± 0.40 & \textbf{82.40 ± 0.17} & 53.65 ± 0.28 & \textbf{93.16 ± 0.22} & \textbf{27.27 ± 0.75} & \textbf{89.51} & 35.06 \\
ASH~\cite{ash} & 78.49 ± 2.58 & 79.03 ± 4.22 & 80.58 ± 0.66 & 59.20 ± 2.46 & \textbf{93.90 ± 0.27} & \textbf{27.29 ± 1.12} & 84.32 & 55.17 \\
GEN~\cite{gen} & 91.35 ± 0.69 & 34.73 ± 1.58 & 79.68 ± 0.75 & 56.71 ± 1.59 & 91.36 ± 0.10 & 32.10 ± 0.59 & 87.46 & 41.18 \\
ExCeL~\cite{excel} & 91.69 ± 0.18 & 40.03 ± 0.84 & \textbf{82.04 ± 0.90} & \textbf{52.24 ± 1.90} & 91.97 ± 0.27 & 28.45 ± 0.80 & 88.57 & 40.24 \\
\hline
\multicolumn{9}{c}{Training methods without outliers} \\
\hline
\rowcolor{gray!30}
RankOOD (ours) & 93.19 ± 0.84 & 20.96 ± 2.55 & \textbf{83.63 ± 1.06} & \textbf{47.44 ± 0.80} & 92.14 ± 0.20 & \textbf{27.73 ± 0.33} & \textbf{89.65} & \textbf{32.04} \\
\hline
CRAFT~\cite{karunanayake2025craft} & 93.94 ± 0.20 & \textbf{19.40 ± 0.88} & 82.03 ± 0.34 & \textbf{51.86 ± 0.49} & 90.88 ± 0.89 & 32.67 ± 1.13 & \textbf{88.95} & 34.64 \\
ConfBranch~\cite{confbranch} & 92.85 ± 0.29 & 21.48 ± 0.94 & 68.90 ± 1.83 & 71.82 ± 3.39 & 90.43 ± 0.18 & 34.75 ± 0.63 & 84.06 & 42.68 \\
G-ODIN~\cite{godin} & \textbf{95.51 ± 0.31} & 21.45 ± 1.91 & \textbf{85.67 ± 1.58} & \textbf{42.68 ± 3.19} & \textbf{92.33 ± 0.11} & 30.18 ± 0.49 & \textbf{91.17} & \textbf{31.44} \\
LogitNorm~\cite{logitnorm} & \textbf{96.74 ± 0.06} & \textbf{13.81 ± 0.20} & 81.53 ± 1.26 & 53.61 ± 3.45 & \textbf{93.04 ± 0.21} & \textbf{26.11 ± 0.52} & \textbf{90.44} & \textbf{31.18} \\
CIDER~\cite{cider} & \textbf{94.71 ± 0.36} & \textbf{20.72 ± 0.85} & 80.49 ± 0.68 & 54.22 ± 1.24 & 90.66 ± 1.68 & 30.17 ± 2.75 & 88.62 & 35.04 \\
\hline
\multicolumn{9}{c}{Training methods with outliers}  \\
\hline
OE~\cite{oe} & \textbf{96.00 ± 0.13} & \textbf{13.13 ± 0.53} & 81.41 ± 1.49 & 54.82 ± 2.79 & 89.02 ± 0.18 & 34.17 ± 0.56 & 88.81 & \textbf{34.04} \\
MCD~\cite{mcd} & 91.00 ± 1.10 & 32.03 ± 4.21 & 74.72 ± 0.78 & 54.39 ± 1.34 & 88.94 ± 0.10 & 29.93 ± 0.30 & 84.89 & 38.78 \\
UDG~\cite{udg} & \textbf{94.06 ± 0.90} & \textbf{20.35 ± 2.41} & 79.59 ± 1.77 & 59.00 ± 3.35 & 82.09 ± 2.78 & 62.04 ± 5.99 & 85.25 & 47.13 \\
MixOE~\cite{mixoe} & 91.93 ± 0.69 & 33.84 ± 4.77 & 76.40 ± 1.44 & 63.88 ± 2.48 & 88.27 ± 0.41 & 40.93 ± 0.29 & 85.53 & 46.22 \\
\hline
\end{tabular}
\end{table*}

\subsection{Qualitative Study of Rank Distributions}
\begin{figure*}[t]
    \centering
    \includegraphics[width=0.99\linewidth]{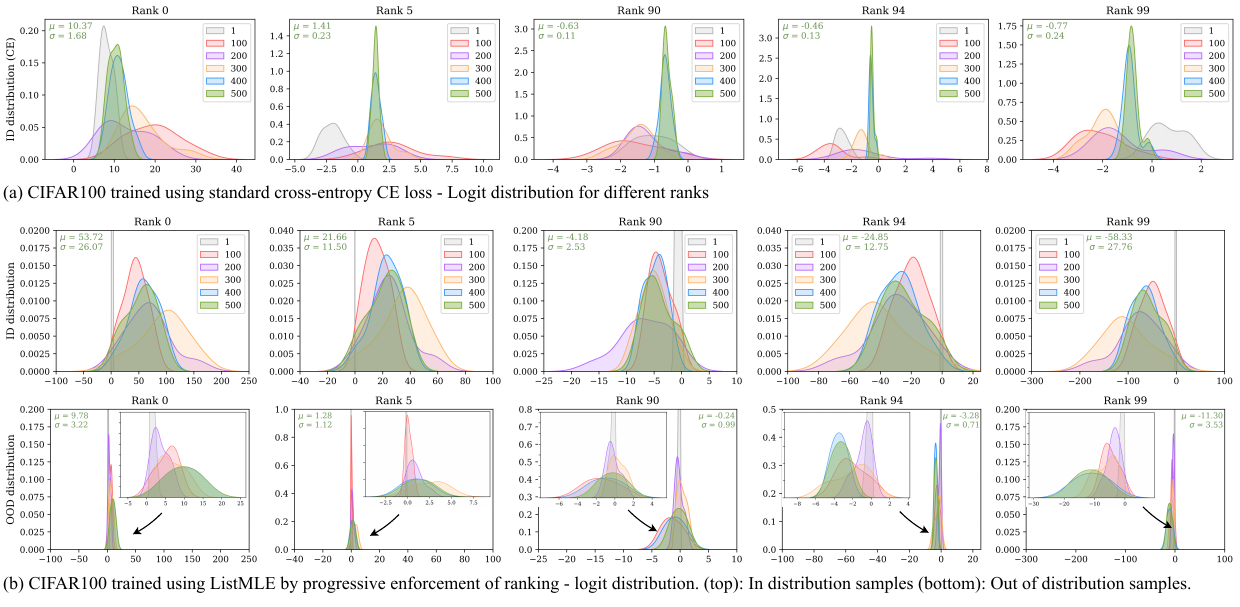}
    \caption{Logit distributions at selected rank positions for predicted class $82$ on CIFAR100. \textit{Across training epochs, (b)~ID samples show increasing separation across rank positions, while (b)~OOD samples remain concentrated near zero with minimal separation.} OOD samples are shown for comparison only.}
    \label{fig: logit_rank_distributions}
    \vspace{-4mm}
\end{figure*}

To analyze how rank-based training influences logit distributions, we examine the class-conditional logit distributions across different rank positions on CIFAR-100 under (a) standard CE training and (b) RankOOD-T. For a given predicted class, we sample logits corresponding to five rank positions (rank-0, three intermediate ranks, and rank-99) and measure their evolution across training epochs. Next, we compare these distributions for ID and OOD samples. 

As shown in Fig.~\ref{fig: logit_rank_distributions}(b), after convergence, the mean logits at ranks $0$, $90$, and $99$ are approximately $53.72$, $-4.18$, and $-58.33$, respectively, yielding a consistent minimal average rank separation margin of roughly $25$ (difference of mean values between distributions). This is because ListMLE optimizes the pairwise logit ordering likelihood and therefore depends only on the relative margin between logits of adjacent ranks, penalizing margin violations more strongly than large positive margins~\cite{xia2008listwise}. Consequently, the optimization explicitly prioritizes preserving the correct ordering rather than pushing the correct class logit arbitrarily high. ListMLE progressively induces a structured semantic hierarchy along the rank dimension: logits associated with semantically closer classes remain comparatively high, while less similar classes are increasingly suppressed, leading to a stable and widening margin across the ranking spectrum. In contrast, low ranks (semantically nearest neighbors) in the CE model initially remain relatively higher, their separation decreases as training progresses, leading to substantial overlap in the logit distributions for mid- and high-rank classes and losing semantic ordering information. Moreover, as shown in Fig.~\ref{fig: logit_rank_distributions}(a), CE induces rank distribution compression, where semantically dissimilar classes become difficult to distinguish in later rank positions.

In the bottom row of Fig.~\ref{fig: logit_rank_distributions}(b), we report the rank-wise logit distributions for OOD samples. Unlike ID samples, OOD logits cluster near zero with substantially lower variance, forming a compact distribution. Nonetheless, a small yet consistent average margin between ranks remains. This is because ListMLE enforces relative logit margins only for ID samples, which indirectly shapes the geometry of the classifier weights. When an OOD feature lies closer to certain ID class regions, the corresponding logits reflect this proximity, producing slight rank-dependent variation. However, as shown in the Sec.~\ref{sec:logit_ordering_conditional_prob} and Fig.~\ref{fig:conditional_prob}, OOD samples do not maintain expected class ranking order for the predicted class.

Overall, the logit magnitude between the ID and OOD in different ranks enables effective OOD detection: ID samples maintain higher, stable rank-dependent margins, while OOD samples remain near zero across ranks.

\subsection{Logit Ordering Consistency Explains Improved OOD Detection under RankOOD}
\label{sec:logit_ordering_conditional_prob}
\begin{figure}[t]
    \centering
    \includegraphics[width=\linewidth]{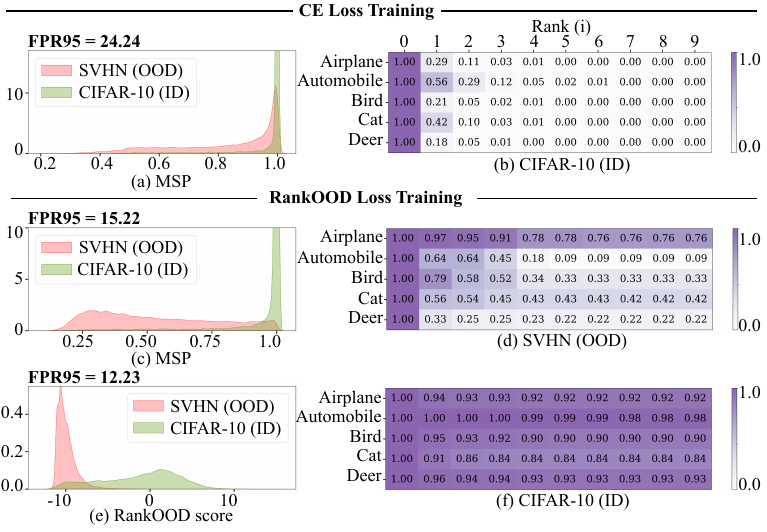}
    \caption{Left: Distributions of RankOOD-S and MSP scores for CIFAR-10 (ID) and SVHN (OOD) samples under RankOOD-T ad CE training. Right: Conditional probability matrix (CP) of rank position $i$ given that all prior ranks have been correctly predicted.}
    \label{fig:conditional_prob}
    \vspace{-6mm}
\end{figure}
The left column of Fig.~\ref{fig:conditional_prob} compares the distributions of MSP and RankOOD-S for CIFAR-10 (ID) and SVHN (OOD) samples under models trained with CE and RankOOD-T losses. Under CE training, MSP (\textbf{cf.} Fig.~\ref{fig:conditional_prob}(a)) achieves an FPR95 of 24.24 on SVHN, while RankOOD reduces it by $37.12\%$ (\textbf{cf.} Fig.~\ref{fig:conditional_prob}(c)). The RankOOD-T loss explicitly enforces a consistent ranking structure among logits through ListMLE supervision. This encourages the logits to maintain smooth, monotonic decay rather than allowing a single confidence spike. Consequently, even when an OOD input produces a high top-1 logit, its intermediate logits tend to deviate from the canonical decay pattern observed in ID data. This structural deviation reduces the overall MSP score, shifting OOD MSP scores away from $1$ and reducing FPR95 for MSP under RankOOD-T. Moreover, the RankOOD-S further decreases FPR95 to $12.23$ (Fig.~\ref{fig:conditional_prob}(e)). This occurs because the RankOOD-S amplifies this effect by penalizing rank-order violations and deviations from class-specific logit thresholds.

To analyze how RankOOD leverages ranking consistency, the right column shows conditional probability (CP) matrices capturing $P(rank~ i \mid ranks<i ~are~ correct)$. We report three cases: (b) CIFAR-10 (ID) under CE training, (d) SVHN (OOD) under RankOOD-T, and (f) CIFAR-10 (ID) under RankOOD-T. Due to space constraints, we report CP matrices for five classes, and provide the complete figures in the Appendix~A.6. 
Each CP matrix illustrates the model’s ability to preserve the class-wise canonical rank order across ranks. While the CE-trained model ({\bf cf.} Fig.~\ref{fig:conditional_prob}(b)) shows weak rank-consistency, RankOOD-T induces sequential dependencies between ranks, yielding high conditional probabilities $(>0.84)$ for ID data ({\bf cf.} Fig.~\ref{fig:conditional_prob}(f)). In contrast, OOD samples under RankOOD-T loss exhibit early rank-order violations, accumulating higher penalties $\delta_{x_{i}}$, in Eq.~\ref{eq:penalty} and causing higher deviation from the reference logit thresholds $Ref_{i}$. These deviations contribute to RankOOD’s superior sensitivity to OOD distortions.

\subsection{Ablation Study}
\label{sec:ablation}
\noindent\textbf{Subset of Ranks:}
\begin{figure}[t]
    \centering
    \includegraphics[width=\linewidth]{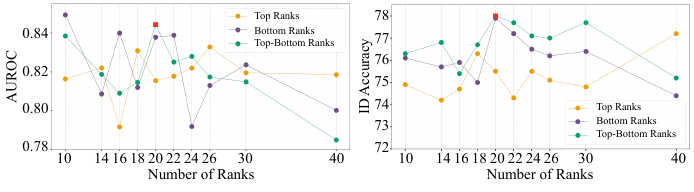}
    \caption{OOD and ID detection performance on CIFAR-100 when training with different rank subsets. (left): AUROC (right): ID accuracy. Top Ranks uses the top$-N$ class ranks, Bottom Ranks uses the lowest $N-1$ ranks along with rank$-0$, and Top-Bottom samples top-N/2 and lowest N/2 ranks.}
    \label{fig:rank_selection}
    \vspace{-5mm}
\end{figure}
In Fig.~\ref{fig:rank_selection} (left), we evaluate OOD detection on CIFAR-100 when training with three types of rank subsets; Top ranks subset uses the top-N class ranks, Bottom ranks subset uses the lowest N-1 ranks with rank-0 and Top-Bottom combines the top-N/2 and botom-N/2 class ranks, respectively. 
Since the selection of rankings is not only crucial for OOD detection performance but also for ID accuracy, we illustrate the right sub-figure according to the same three subsets. As shown, using only the Top ranks yields lower ID accuracy since it focuses on separating semantically similar classes. In contrast, including Bottom or both Top–Bottom ranks improve ID accuracy by exposing the model to semantically dissimilar classes. Both the top and Bottom rank subsets exhibit unstable AUROC when trained with a small number of ranks. This behavior arises because the Top ranks primarily distinguish semantically similar class variations, leading to dominance of near-OOD samples, whereas the Bottom ranks, in high-level learns to differentiate highly dissimilar classes compared to rank-0 via logit separations, reducing sensitivity to near-OOD regions.
When selecting Top-Bottom ranks, after a certain point, the performance decreases as the ranks introduce more noise. Empirically, selecting the top $10$ and bottom $10$ ranks ($N{=}20$) achieves the best balance - maximizing ID accuracy and maintaining high AUROC. This shows that enforcing Listwise ranking across semantically diverse ranks enhances OOD detection. We observed these numbers based on an $\alpha = 1.0$ and a logit threshold of $0.95$.
\\
\noindent\textbf{Ablation of $\alpha$ and Logit threshold:}
\begin{figure}[t]
    \centering
    \includegraphics[width=\linewidth]{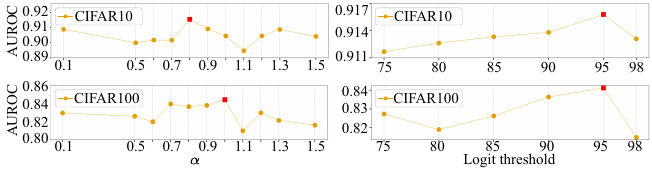}
    \caption{OOD detection performance on CIFAR-100 with respect to $\alpha$ and logit threshold ($Ref$) hyper-parameters.}
    \label{fig:alpha_selection}
    \vspace{-5mm}
\end{figure}
In Fig.~\ref{fig:alpha_selection}, we present the ablation results of selecting the alpha in Eq.~\ref{eq: loss} and the logit threshold ($Ref$) in the RankOOD-S. We conducted AUROC tests with different alpha and logit thresholds. For both datasets, $\alpha=0.1$ shows high AUROC as the loss function is closer to CE loss, and the highest AUROC is achieved at $0.8$ for CIFAR-10 and $1.0$ for CIFAR-100.
With both datasets, we observe AUROC increasing with the increments in the logit threshold, peaking at around 0.95 and starting to decline as the increasing $Ref$ increases the gap between OOD rank logit and threshold in lower ranks, whereas ID samples exhibit a lower gap and vice versa for higher ranks. Consequently, the AUROC decreases once $Ref$ surpasses this threshold, as the threshold becomes the maximum logit value of the given rank. This variation demonstrates significant performance improvement.

\noindent\textbf{Effectiveness of RankOOD-T: }We evaluate RankOOD-T with various logit-based OOD scores and consistently outperform standard CE training. These results suggest that primary gains are due to
the RankOOD-T objective
; detailed results are provided in  Appendix~A.3.

\section{Concluding Remarks}
\label{Conclusion_chp4}

In this paper, we presented a novel OOD detection framework, \textit{RankOOD}, which leverages class-wise logit ranking structures to enhance the detection of out-of-distribution samples. RankOOD constructs canonical rank orders using a 0–1 integer linear programming formulation guided by semantic similarity from probability mass functions of a pretrained network, and optimizes them via a Listwise Maximum Likelihood Estimation (ListMLE) objective. Extensive evaluations on CIFAR-10, CIFAR-100, and TinyImageNet demonstrate that RankOOD consistently ranks among the top two methods for near-OOD detection and the top three for far-OOD detection across $34$ baselines. Notably, RankOOD achieves superior AUROC and FPR95 on the challenging TinyImageNet near-OOD benchmark, highlighting its ability to capture fine-grained semantic distinctions between ID and OOD samples.
{
\small
\bibliographystyle{ieeenat_fullname}
\bibliography{main}
}
\emph{}\\\\
\section*{ Acknowledgment} 
This research was supported by the Australian Research Council’s Discovery Project's funding scheme (Project ID DP220102520).

\clearpage
\appendix
\setcounter{page}{1}
\maketitlesupplementary

\section{Introduction}

In this supplementary material, we present additional details and results that were excluded from the main content due to space limitations. 
\begin{table}[b]
\centering
\footnotesize
\caption{Per-epoch GPU time (sec.) and per-class ILP runtime (sec.) across different datasets (C-CIFAR, IN-ImageNet).}

\label{tab:gpu_time}
\renewcommand{\arraystretch}{0.9}
\setlength{\tabcolsep}{4pt}
\begin{tabular}{lccccc}
\toprule 
\textbf{Method} & \textbf{C-10} & \textbf{C-100} & \textbf{IN-200} & \textbf{IN-1k} & \textbf{Epochs} \\
\midrule
LogitNorm & 8.33  & 5.32  & 1045.50 & 6614.0 & $\sim200$\\
CRAFT     & 17.45 & 25.44 & 1354.00 & - & $\sim100$\\
RankOOD   & 13.80  & 14.34 & 1402.12 & 6855.7 & $\sim300-500$\\
\midrule
ILP Runtime & 0.0058 & 0.1718 & 2.4933 & 1413.6 & - \\
ILP - Greedy & 0.0006 & 0.0023 & 0.0177 & 0.2147 & - \\
\bottomrule
\vspace{-7mm}
\end{tabular}
\end{table}

\subsection{Comparison with baselines}
\label{sec:appendix_baseline_comp}

In the main text (Tab.~2 and Tab.~3)
, we report results only for baselines that placed in the top five in at least one setting. For completeness, we provide the full results for all 35 baselines in Tab.~\ref{tab:nearood_appendix} (near-OOD) and Tab.~\ref{tab:farood_appendix} (far-OOD), with key observations summarized in the respective captions.

\begin{table*}[t!]
\scriptsize
\centering
\captionsetup{width=0.85\linewidth}
\caption{ Performance comparison in \textit{near-OOD detection}. For each column, the top five methods are marked in \textbf{bold}. Note that N/A indicates missing results in OpenOOD. 
\textit{OE obtains the lowest average FPR95 (34.29), while RankOOD ranks second with 44.79. However, OE’s performance is known to be biased toward seen outliers used during training, making its evaluation less reliable~\cite{karunanayake2025craft}. Among methods that do not rely on outliers, RankOOD achieves the best performance for both AUROC and FPR95. Notably, RankOOD achieves state-of-the-art performance on TinyImageNet near-OOD, reducing FPR95 by 4.3\% relative to the strongest baseline OE. }}
\label{tab:nearood_appendix}
\begin{tabular}{lcc|cc|cc|cc}
\hline
Method & \multicolumn{2}{c|}{CIFAR-10} & \multicolumn{2}{c|}{CIFAR-100} & \multicolumn{2}{c|}{TinyImageNet} & \multicolumn{2}{c}{Average} \\
 & AUROC $\uparrow$ & FPR95 $\downarrow$ & AUROC $\uparrow$ & FPR95 $\downarrow$ & AUROC $\uparrow$ & FPR95 $\downarrow$ & AUROC $\uparrow$ & FPR95 $\downarrow$ \\
\hline
\multicolumn{9}{c}{Post-hoc inference methods} \\ 
\hline
OpenMax~\cite{openmax} & 87.62 ± 0.29 & 43.62 ± 2.27 & 76.41 ± 0.25 & 56.58 ± 0.73 & 80.27 ± 0.10 & 63.48 ± 0.25  & 81.43 & 54.56 \\
MSP~\cite{msp} & 88.03 ± 0.25 & 48.17 ± 3.92 & 80.27 ± 0.11 & \textbf{54.80 ± 0.33} & 83.34 ± 0.06 & 54.82 ± 0.35 & 83.88 & 52.60 \\
TempScale~\cite{tempscale} & 88.09 ± 0.31 & 50.96 ± 4.32 & 80.90 ± 0.07 & \textbf{54.49 ± 0.48} & \textbf{83.69 ± 0.04} & 54.82 ± 0.23 & 84.23 & 53.42 \\
ODIN~\cite{odin} & 82.87 ± 1.85 & 76.19 ± 6.08 & 79.90 ± 0.11 & 57.91 ± 0.51 & 80.27 ± 0.08 & 66.76 ± 0.26 & 81.01 & 66.95 \\
MDS~\cite{mds} & 84.20 ± 2.40 & 49.90 ± 3.98 & 58.69 ± 0.09 & 83.53 ± 0.60 & 61.93 ± 0.51 & 79.11 ± 0.31  & 68.27 & 70.85 \\
MDSEns~\cite{mds} & 60.43 ± 0.26 & 92.26 ± 0.20 & 46.31 ± 0.24 & 95.88 ± 0.04 & 54.32 ± 0.24 & 91.75 ± 0.10 & 53.69 & 93.30 \\
RMDS~\cite{rmds} & 89.80 ± 0.28 & 38.89 ± 2.39 & 80.15 ± 0.11 & 55.46 ± 0.41 & 82.57 ± 0.25 & \textbf{54.02 ± 0.58} & 84.17 & \textbf{49.46} \\
Gram~\cite{gram} & 58.66 ± 4.83 & 90.87 ± 1.91 & 51.66 ± 0.77 & 92.28 ± 0.29 & 67.67 ± 1.07 & 86.40 ± 1.21 & 59.33 & 89.85 \\
EBO~\cite{ebo} & 87.58 ± 0.46 & 61.34 ± 4.63 & \textbf{80.91 ± 0.08} & 55.62 ± 0.61 & 82.50 ± 0.05 & 60.24 ± 0.57 & 83.66 & 59.07 \\
OpenGAN~\cite{opengan} & 53.71 ± 7.68 & 94.48 ± 4.01 & 65.98 ± 1.26 & 76.52 ± 2.59 & 59.79 ± 3.39 & 84.15 ± 3.85 & 59.83 & 85.05 \\
GradNorm~\cite{gradnorm} & 54.90 ± 0.98 & 94.72 ± 0.82 & 70.13 ± 0.47 & 85.58 ± 0.46 & 72.75 ± 0.48 & 82.67 ± 0.30 & 65.93 & 87.66 \\
ReAct~\cite{react} & 87.11 ± 0.61 & 63.56 ± 7.33 & 80.77 ± 0.05 & 56.39 ± 0.34 & 81.87 ± 0.98 & 62.49 ± 2.19 & 83.25 & 60.81 \\
MLS~\cite{mls} & 87.52 ± 0.47 & 61.32 ± 4.62 & \textbf{81.05 ± 0.07} & 55.47 ± 0.66 & 82.90 ± 0.04 & 59.76 ± 0.59 & 83.82 & 58.85 \\
KLM~\cite{mls} & 79.19 ± 0.80 & 87.86 ± 6.37 & 76.56 ± 0.25 & 77.92 ± 1.31 & 80.76 ± 0.08 & 70.26 ± 0.64 & 78.84 & 78.68 \\
VIM~\cite{vim} & 88.68 ± 0.28 & 44.84 ± 2.31 & 74.98 ± 0.13 & 62.63 ± 0.27 & 78.68 ± 0.24 & 59.19 ± 0.71 & 80.78 & 55.55 \\
KNN~\cite{knn} & 90.64 ± 0.20 & 34.01 ± 0.38 & 80.18 ± 0.15 & 61.22 ± 0.14 & 81.57 ± 0.17 & 60.18 ± 0.52 & 84.13 & 51.80 \\
DICE~\cite{dice} & 78.34 ± 0.79 & 70.04 ± 7.64 & 79.38 ± 0.23 & 57.95 ± 0.53 & 81.78 ± 0.14 & 61.88 ± 0.67 & 79.83 & 63.29 \\
RankFeat~\cite{rankfeat} & 79.46 ± 2.52 & 60.88 ± 4.60 & 61.88 ± 1.28 & 80.59 ± 1.10 & 56.92 ± 1.59 & 92.06 ± 0.23 & 66.09 & 77.84 \\
ASH~\cite{ash} & 75.27 ± 1.04 & 86.78 ± 1.82 & 78.20 ± 0.15 & 65.71 ± 0.24 & 82.38 ± 0.19 & 64.89 ± 0.90 & 78.62 & 72.46 \\
SHE~\cite{she} & 81.54 ± 0.51 & 79.65 ± 3.47 & 78.95 ± 0.18 & 59.07 ± 0.25 & 80.18 ± 0.25 & 66.80 ± 0.74 & 80.22 & 68.51 \\
GEN~\cite{gen} & 88.20 ± 0.30 & 53.67 ± 3.14 & \textbf{81.31 ± 0.08} & \textbf{54.42 ± 0.33} & \textbf{83.68 ± 0.06} & 55.20 ± 0.20 & \textbf{84.40} & 54.43 \\
ExCeL~\cite{excel} & 86.89 ± 0.23 & 66.55 ± 0.43 & 80.70 ± 0.06 & 55.21 ± 0.56 & 82.40 ± 0.04 & 57.90 ± 0.40 & 83.33 & 59.89 \\
\hline
\multicolumn{9}{c}{Training methods without outliers} \\
\hline
\rowcolor{gray!30}
RankOOD (ours) & 90.21 ± 0.41 & \textbf{31.72 ± 0.67} & 80.67 ± 0.40 & \textbf{52.59 ± 0.75} & \textbf{85.30 ± 0.18} & \textbf{50.05 ± 0.16} & \textbf{85.39} & \textbf{44.79} \\
\hline
CRAFT\cite{karunanayake2025craft} & \textbf{91.11 ± 0.04} & 31.94 ± 1.41 & 80.90 ± 0.33 & 53.73 ± 0.62 & \textbf{83.65 ± 0.41} & \textbf{54.62 ± 0.57} & \textbf{85.22} & 46.76\\

ConfBranch~\cite{confbranch} & 89.84 ± 0.24 & \textbf{31.28 ± 0.66} & 71.60 ± 0.62 & 70.21 ± 0.83 & 79.10 ± 0.24 & 61.44 ± 0.34 & 80.18 & 54.31 \\
G-ODIN~\cite{godin} & 89.12 ± 0.57 & 45.54 ± 2.52 & 77.15 ± 0.28 & 67.58 ± 0.98 & 77.28 ± 0.10 & 69.87 ± 0.46 & 81.18 & 61.00 \\
CSI~\cite{csi} & 89.51 ± 0.19 & 33.66 ± 0.64 & 71.45 ± 0.27 & 70.26 ± 0.56 & N/A & N/A & 80.48 & 51.96 \\
ARPL~\cite{arpl} & 87.44 ± 0.15 & 40.33 ± 0.70 & 74.94 ± 0.93 & 61.56 ± 1.81 & 82.02 ± 0.10 & 55.74 ± 0.70 & 81.47 & 52.54 \\
MOS~\cite{mos} & 71.45 ± 3.09 & 78.72 ± 5.86 & 80.40 ± 0.18 & 56.05 ± 1.01 & 69.84 ± 0.46 & 71.60 ± 0.48 & 73.90 & 68.79 \\
LogitNorm~\cite{logitnorm} & \textbf{92.33 ± 0.08} & \textbf{29.34 ± 0.81} & 78.47 ± 0.31 & 62.89 ± 0.57 & 82.66 ± 0.15 & 56.46 ± 0.37 & \textbf{84.49} & \textbf{49.56} \\
CIDER~\cite{cider} & \textbf{90.71 ± 0.16} & 32.11 ± 0.94 & 73.10 ± 0.39 & 72.02 ± 0.31 & 80.58 ± 1.75 & 60.10 ± 0.73 & 81.46 & 54.74 \\
\hline
\multicolumn{9}{c}{Training methods with outliers} \\
\hline
OE~\cite{oe} & \textbf{94.82 ± 0.21} & \textbf{19.84 ± 0.95} & \textbf{88.30 ± 0.10} & \textbf{30.73 ± 0.11} & \textbf{84.84 ± 0.16} & \textbf{52.30 ± 0.67} & \textbf{89.32} & \textbf{34.29} \\
MCD~\cite{mcd} & \textbf{91.03 ± 0.12} & \textbf{30.17 ± 0.06} & 77.07 ± 0.32 & 55.88 ± 0.85 & 83.62 ± 0.09 & \textbf{54.71 ± 0.83} & 83.91 & \textbf{46.92} \\
UDG~\cite{udg} & 89.91 ± 0.25 & 35.34 ± 0.95 & 78.02 ± 0.10 & 61.42 ± 0.48 & 74.30 ± 1.63 & 68.89 ± 1.72 & 80.74 & 55.22 \\
MixOE~\cite{mixoe} & 88.73 ± 0.82 & 51.45 ± 7.78 & \textbf{80.95 ± 0.20} & 55.22 ± 0.49 & 82.62 ± 0.03 & 57.97 ± 0.40 & 84.10 & 54.88 \\
\hline
\end{tabular}
\end{table*}


\begin{table*}[t!]
\scriptsize
\centering
\captionsetup{width=0.85\linewidth}
\caption{ Performance comparison in \textit{far-OOD detection}. For each column, the top five methods are marked in \textbf{bold}. Note that N/A indicates missing results in OpenOOD. \textit{RankOOD ranks third in both AUROC (89.65) and FPR95 (32.04) on far-OOD detection. LogitNorm and G-ODIN achieve the strongest results, outperforming RankOOD by roughly 2.7\% in FPR95. Nonetheless, RankOOD surpasses G-ODIN on CIFAR-10 and TinyImageNet and exceeds LogitNorm on CIFAR-100. Moreover, RankOOD outperforms OE on CIFAR-100 and TinyImageNet, while achieving an overall AUROC within 1\% of the best-performing outlier based methods.}}
\label{tab:farood_appendix}
\begin{tabular}{lcc|cc|cc|cc}
\hline
Method & \multicolumn{2}{c|}{CIFAR-10} & \multicolumn{2}{c|}{CIFAR-100} & \multicolumn{2}{c|}{TinyImageNet} & \multicolumn{2}{c}{Average} \\
 & AUROC $\uparrow$ & FPR95 $\downarrow$ & AUROC $\uparrow$ & FPR95 $\downarrow$ & AUROC $\uparrow$ & FPR95 $\downarrow$ & AUROC $\uparrow$ & FPR95 $\downarrow$ \\
\hline
\multicolumn{9}{c}{Post-hoc inference methods} \\
\hline
OpenMax~\cite{openmax} & 89.62 ± 0.19 & 29.69 ± 1.21 & 79.48 ± 0.41 & 54.50 ± 0.68 & 90.20 ± 0.17 & 33.12 ± 0.66 & 86.43 & 39.10 \\
MSP~\cite{msp} & 90.73 ± 0.43 & 31.72 ± 1.84 & 77.76 ± 0.44 & 58.70 ± 1.06 & 90.13 ± 0.09 & 35.43 ± 0.38 & 86.21 & 41.95 \\
TempScale~\cite{tempscale} & 90.97 ± 0.52 & 33.48 ± 2.39 & 78.74 ± 0.51 & 57.94 ± 1.14 & 90.82 ± 0.09 & 34.00 ± 0.37 & 86.84 & 41.81 \\
ODIN~\cite{odin} & 87.96 ± 0.61 & 57.62 ± 4.24 & 79.28 ± 0.21 & 58.86 ± 0.79 & 91.71 ± 0.19 & 34.23 ± 1.05 & 86.32 & 50.24 \\
MDS~\cite{mds} & 89.72 ± 1.36 & 32.22 ± 3.40 & 69.39 ± 1.39 & 72.26 ± 1.56 & 74.72 ± 0.26 & 61.66 ± 0.27 & 77.94 & 55.38 \\
MDSEns~\cite{mds} & 73.90 ± 0.27 & 61.47 ± 0.48 & 66.00 ± 0.69 & 66.74 ± 1.04 & 69.27 ± 0.57 & 80.96 ± 0.38 & 69.72 & 69.72 \\
RMDS~\cite{rmds} & 92.20 ± 0.21 & 25.35 ± 0.73 & \textbf{82.92 ± 0.42} & 52.81 ± 0.63 & 88.06 ± 0.34 & 32.45 ± 0.79 & 87.73 & 36.87 \\
Gram~\cite{rmds} & 71.73 ± 3.20 & 72.34 ± 6.73 & 73.36 ± 1.08 & 64.44 ± 2.37 & 71.19 ± 0.24 & 84.36 ± 0.78 & 72.09 & 73.71 \\
EBO~\cite{ebo} & 91.21 ± 0.92 & 41.69 ± 5.32 & 79.77 ± 0.61 & 56.59 ± 1.38 & 90.86 ± 0.21 & 34.86 ± 1.30 & 87.28 & 44.38 \\
OpenGAN~\cite{opengan} & 54.61 ± 15.5 & 83.52 ± 11.63 & 67.88 ± 7.16 & 70.49 ± 7.38 & 73.15 ± 4.07 & 64.16 ± 9.33 & 65.21 & 72.72 \\
GradNorm~\cite{gradnorm} & 57.55 ± 3.22 & 91.90 ± 2.23 & 69.14 ± 1.05 & 83.68 ± 1.92 & 84.26 ± 0.87 & 66.45 ± 0.22 & 70.32 & 80.68 \\
ReAct~\cite{react} & 90.42 ± 1.41 & 44.90 ± 8.37 & 80.39 ± 0.49 & 54.20 ± 1.56 & \textbf{92.31 ± 0.56} & 28.50 ± 0.95 & 87.71 & 42.53 \\
MLS~\cite{mls} & 91.10 ± 0.89 & 41.68 ± 5.27 & 79.67 ± 0.57 & 56.73 ± 1.33 & 91.11 ± 0.19 & 34.03 ± 1.21 & 87.29 & 44.15 \\
KLM~\cite{mls} & 82.68 ± 0.21 & 78.31 ± 4.84 & 76.24 ± 0.52 & 71.65 ± 2.01 & 88.53 ± 0.11 & 40.90 ± 1.08 & 82.48 & 63.62 \\
VIM~\cite{vim} & 93.48 ± 0.24 & 25.05 ± 0.52 & 81.70 ± 0.62 & \textbf{50.74 ± 1.00} & 91.26 ± 0.19 & \textbf{27.20 ± 0.30} & \textbf{88.81} & \textbf{34.33} \\
KNN~\cite{knn} & 92.96 ± 0.14 & 24.27 ± 0.40 & \textbf{82.40 ± 0.17} & 53.65 ± 0.28 & \textbf{93.16 ± 0.22} & \textbf{27.27 ± 0.75} & \textbf{89.51} & 35.06 \\
DICE~\cite{dice} & 84.23 ± 1.89 & 51.76 ± 4.42 & 80.01 ± 0.18 & 56.25 ± 0.60 & 90.80 ± 0.31 & 36.51 ± 1.18 & 85.01 & 48.17 \\
RankFeat~\cite{rankfeat} & 75.87 ± 5.06 & 57.44 ± 7.99 & 67.10 ± 1.42 & 69.45 ± 1.01 & 38.22 ± 3.85 & 97.72 ± 0.75 & 60.40 & 74.87 \\
ASH~\cite{ash} & 78.49 ± 2.58 & 79.03 ± 4.22 & 80.58 ± 0.66 & 59.20 ± 2.46 & \textbf{93.90 ± 0.27} & \textbf{27.29 ± 1.12} & 84.32 & 55.17 \\
SHE~\cite{she} & 85.32 ± 1.43 & 66.48 ± 5.98 & 76.92 ± 1.16 & 64.12 ± 2.70 & 89.81 ± 0.61 & 42.17 ± 1.24 & 84.02 & 57.59 \\
GEN~\cite{gen} & 91.35 ± 0.69 & 34.73 ± 1.58 & 79.68 ± 0.75 & 56.71 ± 1.59 & 91.36 ± 0.10 & 32.10 ± 0.59 & 87.46 & 41.18 \\
ExCeL~\cite{excel} & 91.69 ± 0.18 & 40.03 ± 0.84 & \textbf{82.04 ± 0.90} & \textbf{52.24 ± 1.90} & 91.97 ± 0.27 & 28.45 ± 0.80 & 88.57 & 40.24 \\
\hline
\multicolumn{9}{c}{Training methods without outliers} \\
\hline
\rowcolor{gray!30}
RankOOD (ours) & 93.19 ± 0.84 & 20.96 ± 2.55 & \textbf{83.63 ± 1.06} & \textbf{47.44 ± 0.80} & 92.14 ± 0.20 & \textbf{27.73 ± 0.33} & \textbf{89.65} & \textbf{32.04} \\
\hline
CRAFT~\cite{karunanayake2025craft} & 93.94 ± 0.20 & \textbf{19.40 ± 0.88} & 82.03 ± 0.34 & \textbf{51.86 ± 0.49} & 90.88 ± 0.89 & 32.67 ± 1.13 & \textbf{88.95} & 34.64 \\
ConfBranch~\cite{confbranch} & 92.85 ± 0.29 & 21.48 ± 0.94 & 68.90 ± 1.83 & 71.82 ± 3.39 & 90.43 ± 0.18 & 34.75 ± 0.63 & 84.06 & 42.68 \\
G-ODIN~\cite{godin} & \textbf{95.51 ± 0.31} & 21.45 ± 1.91 & \textbf{85.67 ± 1.58} & \textbf{42.68 ± 3.19} & \textbf{92.33 ± 0.11} & 30.18 ± 0.49 & \textbf{91.17} & \textbf{31.44} \\
CSI~\cite{csi} & 92.00 ± 0.30 & 26.42 ± 0.29 & 66.31 ± 1.21 & 76.92 ± 1.29 & N/A & N/A & 79.16 & 51.67 \\
ARPL~\cite{arpl} & 89.31 ± 0.32 & 32.39 ± 0.74 & 73.69 ± 1.80 & 63.14 ± 2.53 & 89.23 ± 0.11 & 36.46 ± 0.08 & 84.08 & 44.00 \\
MOS~\cite{mos} & 76.41 ± 5.93 & 62.90 ± 6.62 & 80.17 ± 1.21 & 57.28 ± 3.29 & 80.46 ± 0.92 & 51.56 ± 0.42 & 79.01 & 57.25 \\
LogitNorm~\cite{logitnorm} & \textbf{96.74 ± 0.06} & \textbf{13.81 ± 0.20} & 81.53 ± 1.26 & 53.61 ± 3.45 & \textbf{93.04 ± 0.21} & \textbf{26.11 ± 0.52} & \textbf{90.44} & \textbf{31.18} \\
CIDER~\cite{cider} & \textbf{94.71 ± 0.36} & \textbf{20.72 ± 0.85} & 80.49 ± 0.68 & 54.22 ± 1.24 & 90.66 ± 1.68 & 30.17 ± 2.75 & 88.62 & 35.04 \\
\hline
\multicolumn{9}{c}{Training methods with outliers}  \\
\hline
OE~\cite{oe} & \textbf{96.00 ± 0.13} & \textbf{13.13 ± 0.53} & 81.41 ± 1.49 & 54.82 ± 2.79 & 89.02 ± 0.18 & 34.17 ± 0.56 & 88.81 & \textbf{34.04} \\
MCD~\cite{mcd} & 91.00 ± 1.10 & 32.03 ± 4.21 & 74.72 ± 0.78 & 54.39 ± 1.34 & 88.94 ± 0.10 & 29.93 ± 0.30 & 84.89 & 38.78 \\
UDG~\cite{udg} & \textbf{94.06 ± 0.90} & \textbf{20.35 ± 2.41} & 79.59 ± 1.77 & 59.00 ± 3.35 & 82.09 ± 2.78 & 62.04 ± 5.99 & 85.25 & 47.13 \\
MixOE~\cite{mixoe} & 91.93 ± 0.69 & 33.84 ± 4.77 & 76.40 ± 1.44 & 63.88 ± 2.48 & 88.27 ± 0.41 & 40.93 ± 0.29 & 85.53 & 46.22 \\
\hline
\end{tabular}
\end{table*}

\subsection{Detailed per-dataset OOD detection results}
\label{sec:appendix_perdata_results}

In Sec.~5.1
, we reported average AUROC and FPR95 across all OOD datasets for each ID dataset (Tab.~2 and Tab.~3)
. For completeness, we provide the per-dataset results for each OOD benchmark. TinyImageNet results in Tab.~\ref{tab:fpr_tin} and Tab.~\ref{tab:auroc_tin}; CIFAR-100 results in Tab.~\ref{tab:fpr_cifar100} and Tab.~\ref{tab:auroc_cifar100}; and CIFAR-10 results are shown in Tab.~\ref{tab:fpr_cifar10} and Tab.~\ref{tab:auroc_cifar10}.


\begin{table*}[h]
\centering
\scriptsize
\captionsetup{width=0.8\linewidth}
\caption{FPR95 (\% $\downarrow$) of various methods for different OOD datasets when TinyImageNet is ID. \break For each column, the top five methods are marked in \textbf{bold}. Note that N/A indicates that results are not reported in OpenOOD. \textit{RankOOD achieves SOTA performance in near-OOD and ranks within the top three methods in two out of three far-OOD datasets.}}
\label{tab:fpr_tin}
\begin{tabular}{l|ccc|cccc}
\hline
Method & \multicolumn{3}{c|}{Near OOD} & \multicolumn{4}{c}{Far OOD} \\
 & SSB-hard & NINCO & Average & iNaturalist & Textures & OpenImage-O  & Average\\ 
 \hline
 \multicolumn{8}{c}{Post-hoc inference methods} \\ 
\hline
OpenMax~\cite{openmax} & 72.37 ± 0.11 & 54.59 ± 0.54 & 63.48 ± 0.25 & 24.53 ± 0.96 & 36.80 ± 0.55 & 38.03 ± 0.49 & 33.12 ± 0.66 \\
MSP~\cite{msp} & 66.00 ± 0.10 & 43.65 ± 0.75 & 54.82 ± 0.35 & 26.48 ± 0.73 & 44.58 ± 0.68 & 35.23 ± 0.18 & 35.43 ± 0.38 \\
TempScale~\cite{tempscale} & 66.43 ± 0.26 & \textbf{43.21 ± 0.70} & 54.82 ± 0.23 & 24.39 ± 0.79 & 43.57 ± 0.77 & 34.04 ± 0.31 & 34.00 ± 0.37 \\
ODIN~\cite{odin} & 73.51 ± 0.38 & 60.00 ± 0.80 & 66.76 ± 0.26 & 22.39 ± 1.87 & 42.99 ± 1.56 & 37.30 ± 0.59 & 34.23 ± 1.05 \\
MDS~\cite{mds} & 83.65 ± 0.47 & 74.57 ± 0.15 & 79.11 ± 0.31 & 58.53 ± 0.75 & 58.16 ± 0.84 & 68.29 ± 0.28 & 61.66 ± 0.27 \\
MDSEns~\cite{mds} & 92.13 ± 0.05 & 91.36 ± 0.16 & 91.75 ± 0.10 & 83.37 ± 0.70 & 72.27 ± 0.48 & 87.26 ± 0.10 & 80.96 ± 0.38 \\
RMDS~\cite{rmds} & \textbf{65.91 ± 0.27} & \textbf{42.13 ± 1.04} & \textbf{54.02 ± 0.58} & 24.70 ± 0.90 & 37.80 ± 1.32 & 34.85 ± 0.31 & 32.45 ± 0.79 \\
Gram~\cite{gram} & 85.68 ± 0.85 & 87.13 ± 1.89 & 86.40 ± 1.21 & 85.54 ± 0.40 & 80.87 ± 1.20 & 86.66 ± 1.27 & 84.36 ± 0.78 \\
EBO~\cite{ebo} & 69.77 ± 0.32 & 50.70 ± 0.89 & 60.24 ± 0.57 & 26.41 ± 2.29 & 41.43 ± 1.85 & 36.74 ± 1.14 & 34.86 ± 1.30 \\
OpenGAN~\cite{opengan} & 88.07 ± 2.23 & 80.23 ± 5.71 & 84.15 ± 3.85 & 60.13 ± 9.79 & 66.00 ± 9.97 & 66.34 ± 8.44 & 64.16 ± 9.33 \\
GradNorm~\cite{gradnorm} & 82.17 ± 0.62 & 83.17 ± 0.21 & 82.67 ± 0.30 & 61.31 ± 2.86 & 66.88 ± 3.59 & 71.16 ± 0.23 & 66.45 ± 0.22 \\
ReAct~\cite{react} & 71.51 ± 1.92 & 53.47 ± 2.46 & 62.49 ± 2.19 & 22.97 ± 2.25 & \textbf{29.67 ± 1.35} & 32.86 ± 0.74 & 28.50 ± 0.95 \\
MLS~\cite{mls} & 69.64 ± 0.37 & 49.87 ± 0.94 & 59.76 ± 0.59 & 25.09 ± 2.04 & 41.25 ± 1.86 & 35.76 ± 0.74 & 34.03 ± 1.21 \\
KLM~\cite{mls} & 78.19 ± 2.30 & 62.33 ± 2.66 & 70.26 ± 0.64 & 26.66 ± 1.61 & 50.24 ± 1.26 & 45.81 ± 0.59 & 40.90 ± 1.08 \\
VIM~\cite{vim} & 71.28 ± 0.49 & 47.10 ± 1.10 & 59.19 ± 0.71 & 27.34 ± 0.38 & \textbf{20.39 ± 0.17} & 33.86 ± 0.63 & \textbf{27.20 ± 0.30} \\
KNN~\cite{knn} & 73.71 ± 0.31 & 46.64 ± 0.73 & 60.18 ± 0.52 & 24.46 ± 1.06 & \textbf{24.45 ± 0.29} & 32.90 ± 1.12 & \textbf{27.27 ± 0.75} \\
DICE~\cite{dice} & 70.84 ± 0.30 & 52.91 ± 1.20 & 61.88 ± 0.67 & 29.66 ± 2.62 & 40.96 ± 1.87 & 38.91 ± 1.16 & 36.51 ± 1.18 \\
RankFeat~\cite{rankfeat} & 90.79 ± 0.37 & 93.32 ± 0.11 & 92.06 ± 0.23 & 98.00 ± 0.80 & 99.40 ± 0.68 & 95.77 ± 0.85 & 97.72 ± 0.75 \\
ASH~\cite{ash} & 72.14 ± 0.97 & 57.63 ± 0.98 & 64.89 ± 0.90 & 22.49 ± 2.24 & \textbf{25.65 ± 0.80} & 33.72 ± 0.97 & \textbf{27.29 ± 1.12} \\
SHE~\cite{she} & 72.64 ± 0.30 & 60.96 ± 1.33 & 66.80 ± 0.74 & 34.38 ± 3.48 & 45.58 ± 2.42 & 46.54 ± 1.34 & 42.17 ± 1.24 \\
GEN~\cite{gen} & 66.79 ± 0.26 & 43.61 ± 0.61 & 55.20 ± 0.20 & \textbf{22.03 ± 0.98} & 42.01 ± 0.92 & \textbf{32.25 ± 0.31} & 32.10 ± 0.59 \\
ExCeL~\cite{excel} & 69.28 ± 0.60 & 46.51 ± 0.20 & 57.90 ± 0.40 & \textbf{22.29 ± 1.00} & 30.14 ± 0.64 & 32.91 ± 0.76 & 28.45 ± 0.80 \\
\hline
\multicolumn{8}{c}{Training methods without outliers} \\
\hline
\rowcolor{gray!30}
RankOOD (Ours) & \textbf{60.68 ± 0.61} & \textbf{39.43 ± 0.99} & \textbf{50.05 ± 0.16} & \textbf{19.42 ± 0.71} & 33.81 ± 0.95 & \textbf{29.97 ± 0.56} & \textbf{27.73 ± 0.33} \\
\hline
CRAFT ~\cite{karunanayake2025craft} & 69.07 ± 0.39 & \textbf{40.40 ± 1.13} & \textbf{54.62 ± 0.57} & 24.89 ± 1.33 & 38.73 ± 4.55 & 33.48 ± 1.74 & 32.67 ± 1.13 \\
ConfBranch~\cite{confbranch} & 72.24 ± 0.37 & 50.63 ± 0.60 & 61.44 ± 0.34 & 23.84 ± 0.40 & 42.42 ± 2.27 & 37.99 ± 0.09 & 34.75 ± 0.63 \\
G-ODIN~\cite{godin} & 78.23 ± 0.70 & 61.52 ± 0.64 & 69.87 ± 0.46 & 26.13 ± 0.77 & \textbf{28.98 ± 1.15} & 35.43 ± 0.43 & 30.18 ± 0.49 \\
CSI~\cite{csi} & N/A & N/A & N/A & N/A & N/A & N/A & N/A \\
ARPL~\cite{arpl} & \textbf{65.73 ± 0.51} & 45.75 ± 0.89 & 55.74 ± 0.70 & 29.32 ± 0.64 & 42.87 ± 1.09 & 37.20 ± 0.69 & 36.46 ± 0.08 \\
MOS~\cite{mos} & 74.35 ± 0.32 & 68.85 ± 0.68 & 71.60 ± 0.48 & 49.55 ± 0.73 & 51.27 ± 1.02 & 53.86 ± 0.30 & 51.56 ± 0.42 \\
LogitNorm~\cite{logitnorm} & 67.46 ± 0.21 & 45.46 ± 0.69 & 56.46 ± 0.37 & \textbf{15.70 ± 0.61} & 32.13 ± 0.67 & \textbf{30.49 ± 0.62} & \textbf{26.11 ± 0.52} \\
CIDER~\cite{cider} & 75.50 ± 0.68 & 44.69 ± 0.88 & 60.10 ± 0.73 & 26.54 ± 2.27 & 31.51 ± 3.68 & \textbf{32.47 ± 2.40} & 30.17 ± 2.75 \\
\hline
\multicolumn{8}{c}{Training methods with outliers} \\
\hline
OE~\cite{oe} & \textbf{64.67 ± 0.25} & \textbf{39.93 ± 1.13} & \textbf{52.30 ± 0.67} & 27.03 ± 0.47 & 41.92 ± 1.69 & 33.56 ± 0.46 & 34.17 ± 0.56 \\
MCD~\cite{mcd} & \textbf{65.69 ± 0.36} & 43.74 ± 1.32 & \textbf{54.71 ± 0.83} & \textbf{21.74 ± 0.31} & 38.11 ± 0.93 & \textbf{29.93 ± 0.19} & 29.93 ± 0.30 \\
UDG~\cite{udg} & 75.84 ± 1.86 & 61.94 ± 1.61 & 68.89 ± 1.72 & 49.26 ± 7.88 & 71.94 ± 5.25 & 64.92 ± 5.07 & 62.04 ± 5.99 \\
MixOE~\cite{mixoe} & 68.26 ± 0.48 & 47.69 ± 0.95 & 57.97 ± 0.40 & 30.84 ± 0.45 & 51.44 ± 0.61 & 40.51 ± 1.06 & 40.93 ± 0.29 \\
\hline
\end{tabular}
\end{table*}

\begin{table*}[t]
\centering
\scriptsize
\captionsetup{width=0.8\linewidth}
\caption{AUROC (\% $\uparrow$) of various methods for different OOD datasets when TinyImageNet is ID. For each column, the top five methods are marked in \textbf{bold}. Note that N/A indicates that results are not reported in OpenOOD. \textit{RankOOD ranks within  the top three methods in two out of three far-OOD datasets while achieving SOTA performance in near-OOD setting.}}
\label{tab:auroc_tin}
\begin{tabular}{l|ccc|cccc}
\hline
Method & \multicolumn{3}{c|}{Near OOD} & \multicolumn{4}{c}{Far OOD} \\
 & SSB-hard & NINCO & Average & iNaturalist & Textures & OpenImage-O  & Average\\ 
 \hline
 \multicolumn{8}{c}{Post-hoc inference methods} \\ 
\hline
OpenMax~\cite{openmax} & 77.53 ± 0.08 & 83.01 ± 0.17 & 80.27 ± 0.10 & 92.32 ± 0.32 & 90.21 ± 0.07 & 88.07 ± 0.14 & 90.20 ± 0.17 \\
MSP~\cite{msp} & 80.38 ± 0.03 & 86.29 ± 0.11 & 83.34 ± 0.06 & 92.80 ± 0.25 & 88.36 ± 0.13 & 89.24 ± 0.02 & 90.13 ± 0.09 \\
TempScale~\cite{tempscale} & \textbf{80.71 ± 0.02} & \textbf{86.67 ± 0.08} & \textbf{83.69 ± 0.04} & 93.39 ± 0.25 & 89.24 ± 0.11 & 89.84 ± 0.02 & 90.82 ± 0.09 \\
ODIN~\cite{odin} & 77.19 ± 0.06 & 83.34 ± 0.12 & 80.27 ± 0.08 & \textbf{94.37 ± 0.41} & 90.65 ± 0.20 & 90.11 ± 0.15 & 91.71 ± 0.19 \\
MDS~\cite{mds} & 58.38 ± 0.58 & 65.48 ± 0.46 & 61.93 ± 0.51 & 75.03 ± 0.76 & 79.25 ± 0.33 & 69.87 ± 0.14 & 74.72 ± 0.26 \\
MDSEns~\cite{mds} & 50.46 ± 0.36 & 58.18 ± 0.42 & 54.32 ± 0.24 & 62.16 ± 0.73 & 80.70 ± 0.48 & 64.96 ± 0.51 & 69.27 ± 0.57 \\
RMDS~\cite{rmds} & 80.20 ± 0.23 & 84.94 ± 0.28 & 82.57 ± 0.25 & 90.64 ± 0.46 & 86.77 ± 0.38 & 86.77 ± 0.22 & 88.06 ± 0.34 \\
Gram~\cite{gram} & 65.95 ± 1.08 & 69.40 ± 1.07 & 67.67 ± 1.07 & 65.30 ± 0.20 & 80.53 ± 0.37 & 67.72 ± 0.58 & 71.19 ± 0.24 \\
EBO~\cite{ebo} & 79.83 ± 0.02 & 85.17 ± 0.11 & 82.50 ± 0.05 & 92.55 ± 0.50 & 90.79 ± 0.16 & 89.23 ± 0.26 & 90.86 ± 0.21 \\
OpenGAN~\cite{opengan} & 55.08 ± 1.84 & 64.49 ± 4.98 & 59.79 ± 3.39 & 75.32 ± 3.32 & 70.58 ± 4.66 & 73.54 ± 4.48 & 73.15 ± 4.07 \\
GradNorm~\cite{gradnorm} & 72.12 ± 0.43 & 73.39 ± 0.63 & 72.75 ± 0.48 & 86.06 ± 1.90 & 86.07 ± 0.36 & 80.66 ± 1.09 & 84.26 ± 0.87 \\
ReAct~\cite{react} & 78.97 ± 1.33 & 84.76 ± 0.64 & 81.87 ± 0.98 & 93.65 ± 0.88 & \textbf{92.86 ± 0.47} & \textbf{90.40 ± 0.35} & \textbf{92.31 ± 0.56} \\
MLS~\cite{mls} & 80.15 ± 0.01 & 85.65 ± 0.09 & 82.90 ± 0.04 & 93.12 ± 0.45 & 90.60 ± 0.16 & 89.62 ± 0.21 & 91.11 ± 0.19 \\
KLM~\cite{mls} & 77.56 ± 0.18 & 83.96 ± 0.12 & 80.76 ± 0.08 & 91.80 ± 0.21 & 86.13 ± 0.12 & 87.66 ± 0.17 & 88.53 ± 0.11 \\
VIM~\cite{vim} & 74.04 ± 0.31 & 83.32 ± 0.19 & 78.68 ± 0.24 & 90.96 ± 0.36 & \textbf{94.61 ± 0.12} & 88.20 ± 0.18 & 91.26 ± 0.19 \\
KNN~\cite{knn} & 77.03 ± 0.23 & 86.10 ± 0.12 & 81.57 ± 0.17 & \textbf{93.99 ± 0.36} & \textbf{95.29 ± 0.02} & \textbf{90.19 ± 0.32} & \textbf{93.16 ± 0.22} \\
DICE~\cite{dice} & 79.06 ± 0.05 & 84.49 ± 0.24 & 81.78 ± 0.14 & 91.81 ± 0.79 & 91.53 ± 0.21 & 89.06 ± 0.34 & 90.80 ± 0.31 \\
RankFeat~\cite{rankfeat} & 58.74 ± 0.94 & 55.10 ± 2.52 & 56.92 ± 1.59 & 33.08 ± 4.68 & 29.10 ± 2.57 & 52.48 ± 4.44 & 38.22 ± 3.85 \\
ASH~\cite{ash} & 79.52 ± 0.37 & 85.24 ± 0.08 & 82.38 ± 0.19 & \textbf{95.10 ± 0.47} & \textbf{94.77 ± 0.19} & \textbf{91.82 ± 0.25} & \textbf{93.90 ± 0.27} \\
SHE~\cite{she} & 78.30 ± 0.20 & 82.07 ± 0.33 & 80.18 ± 0.25 & 91.43 ± 1.28 & 90.51 ± 0.19 & 87.49 ± 0.70 & 89.81 ± 0.61 \\
GEN~\cite{gen} & \textbf{80.75 ± 0.03} & 86.60 ± 0.08 & \textbf{83.68 ± 0.06} & 93.70 ± 0.18 & 90.25 ± 0.10 & 90.13 ± 0.06 & 91.36 ± 0.10 \\
ExCeL~\cite{excel} & 79.39 ± 0.03 & 85.40 ± 0.04 & 82.40 ± 0.04 & 93.76 ± 0.43 & 92.40 ± 0.05 & 89.75 ± 0.32 & 91.97 ± 0.27 \\
\hline
\multicolumn{8}{c}{Training methods without outliers} \\
\hline
\rowcolor{gray!30}
RankOOD (Ours) & \textbf{82.57 ± 0.28} & \textbf{88.04 ± 0.26} & \textbf{85.30 ± 0.18} & \textbf{94.69 ± 0.17} & 90.85 ± 0.14 & \textbf{90.89 ± 0.25} & 92.14 ± 0.20 \\
\hline
CRAFT~\cite{karunanayake2025craft} & 80.70 ± 0.18 & \textbf{86.74 ± 0.84} & \textbf{83.65 ± 0.41} & 92.85 ± 0.66 & 89.94 ± 0.49 & 89.85 ± 0.46 & 90.88 ± 0.89 \\
ConfBranch~\cite{confbranch} & 75.01 ± 0.35 & 83.19 ± 0.14 & 79.10 ± 0.24 & 93.40 ± 0.09 & 89.64 ± 0.52 & 88.26 ± 0.07 & 90.43 ± 0.18 \\
G-ODIN~\cite{godin} & 72.94 ± 0.05 & 81.63 ± 0.21 & 77.28 ± 0.10 & 93.12 ± 0.21 & \textbf{93.67 ± 0.21} & 90.18 ± 0.15 & \textbf{92.33 ± 0.11} \\
CSI~\cite{csi} & N/A & N/A & N/A & N/A & N/A & N/A & N/A \\
ARPL~\cite{arpl} & 79.24 ± 0.14 & 84.81 ± 0.07 & 82.02 ± 0.10 & 91.54 ± 0.05 & 88.11 ± 0.34 & 88.04 ± 0.20 & 89.23 ± 0.11 \\
MOS~\cite{mos} & 66.54 ± 0.49 & 73.14 ± 0.47 & 69.84 ± 0.46 & 79.69 ± 1.38 & 81.38 ± 0.75 & 80.29 ± 0.68 & 80.46 ± 0.92 \\
LogitNorm~\cite{logitnorm} & 78.42 ± 0.23 & \textbf{86.90 ± 0.07} & 82.66 ± 0.15 & \textbf{96.26 ± 0.20} & 91.85 ± 0.21 & \textbf{91.01 ± 0.27} & \textbf{93.04 ± 0.21} \\
CIDER~\cite{cider} & 76.04 ± 2.37 & 85.13 ± 1.13 & 80.58 ± 1.75 & 90.69 ± 2.13 & 92.38 ± 1.35 & 88.92 ± 1.58 & 90.66 ± 1.68 \\
\hline
\multicolumn{8}{c}{Training methods with outliers} \\
\hline
OE~\cite{oe} & \textbf{82.34 ± 0.16} & \textbf{87.35 ± 0.23} & \textbf{84.84 ± 0.16} & 90.30 ± 0.16 & 87.76 ± 0.32 & 89.01 ± 0.24 & 89.02 ± 0.18 \\
MCD~\cite{mcd} & \textbf{81.51 ± 0.14} & 85.74 ± 0.07 & 83.62 ± 0.09 & 90.83 ± 0.10 & 86.87 ± 0.12 & 89.12 ± 0.18 & 88.94 ± 0.10 \\
UDG~\cite{udg} & 70.73 ± 1.74 & 77.88 ± 1.56 & 74.30 ± 1.63 & 85.95 ± 2.97 & 81.79 ± 2.57 & 78.54 ± 2.98 & 82.09 ± 2.78 \\
MixOE~\cite{mixoe} & 80.23 ± 0.15 & 85.01 ± 0.10 & 82.62 ± 0.03 & 90.64 ± 0.36 & 86.80 ± 0.45 & 87.36 ± 0.49 & 88.27 ± 0.41 \\
\hline
\end{tabular}
\end{table*}


\begin{table*}[h]
\centering
\scriptsize
\captionsetup{width=0.87\linewidth}
\caption{ FPR95 (\% $\downarrow$) of various methods for different OOD datasets when CIFAR-100 is ID. For each column, the top five methods are marked in \textbf{bold}. \textit{RankOOD ranks within the top four methods in four out of six OOD datasets while achieving second best average FPR95 for both near- and far-OOD detection.}}
\label{tab:fpr_cifar100}
\begin{tabular}{l|ccc|ccccc}
\hline
Method & \multicolumn{3}{c|}{Near OOD} & \multicolumn{5}{c}{Far OOD} \\
 & CIFAR-10 & TIN & Average & MNIST & SVHN & Textures & Places365 & Average\\ 
 \hline
 \multicolumn{9}{c}{Post-hoc inference methods} \\ 
\hline
OpenMax~\cite{openmax} & 60.17 ± 0.97 & 52.99 ± 0.51 & 56.58 ± 0.73 & 53.82 ± 4.74 & 53.20 ± 1.78 & 56.12 ± 1.91 & \textbf{54.85 ± 1.42} & 54.50 ± 0.68 \\
MSP~\cite{msp} & \textbf{58.91 ± 0.93} & 50.70 ± 0.34 & \textbf{54.80 ± 0.33} & 57.23 ± 4.68 & 59.07 ± 2.53 & 61.88 ± 1.28 & 56.62 ± 0.87 & 58.70 ± 1.06 \\
TempScale~\cite{tempscale} & \textbf{58.72 ± 0.81} & 50.26 ± 0.16 & \textbf{54.49 ± 0.48} & 56.05 ± 4.61 & 57.71 ± 2.68 & 61.56 ± 1.43 & 56.46 ± 0.94 & 57.94 ± 1.14 \\
ODIN~\cite{odin} & 60.64 ± 0.56 & 55.19 ± 0.57 & 57.91 ± 0.51 & 45.94 ± 3.29 & 67.41 ± 3.88 & 62.37 ± 2.96 & 59.71 ± 0.92 & 58.86 ± 0.79 \\
MDS~\cite{mds} & 88.00 ± 0.49 & 79.05 ± 1.22 & 83.53 ± 0.60 & 71.72 ± 2.94 & 67.21 ± 6.09 & 70.49 ± 2.48 & 79.61 ± 0.34 & 72.26 ± 1.56 \\
MDSEns~\cite{mds} & 95.94 ± 0.16 & 95.82 ± 0.12 & 95.88 ± 0.04 & \textbf{2.83 ± 0.86} & 82.57 ± 2.58 & 84.94 ± 0.83 & 96.61 ± 0.17 & 66.74 ± 1.04 \\
RMDS~\cite{rmds} & 61.37 ± 0.24 & 49.56 ± 0.90 & 55.46 ± 0.41 & 52.05 ± 6.28 & 51.65 ± 3.68 & 53.99 ± 1.06 & \textbf{53.57 ± 0.43} & 52.81 ± 0.63 \\
Gram~\cite{gram} & 92.71 ± 0.64 & 91.85 ± 0.86 & 92.28 ± 0.29 & 53.53 ± 7.45 & \textbf{20.06 ± 1.96} & 89.51 ± 2.54 & 94.67 ± 0.60 & 64.44 ± 2.37 \\
EBO~\cite{ebo} & 59.21 ± 0.75 & 52.03 ± 0.50 & 55.62 ± 0.61 & 52.62 ± 3.83 & 53.62 ± 3.14 & 62.35 ± 2.06 & 57.75 ± 0.86 & 56.59 ± 1.38 \\
OpenGAN~\cite{opengan} & 78.83 ± 3.94 & 74.21 ± 1.25 & 76.52 ± 2.59 & 63.09 ± 23.3 & 70.35 ± 2.06 & 74.77 ± 1.78 & 73.75 ± 8.32 & 70.49 ± 7.38 \\
GradNorm~\cite{gradnorm} & 84.30 ± 0.36 & 86.85 ± 0.62 & 85.58 ± 0.46 & 86.97 ± 1.44 & 69.90 ± 7.94 & 92.51 ± 0.61 & 85.32 ± 0.44 & 83.68 ± 1.92 \\
ReAct~\cite{react} & 61.30 ± 0.43 & 51.47 ± 0.47 & 56.39 ± 0.34 & 56.04 ± 5.66 & 50.41 ± 2.02 & 55.04 ± 0.82 & \textbf{55.30 ± 0.41} & 54.20 ± 1.56 \\
MLS~\cite{mls} & \textbf{59.11 ± 0.64} & 51.83 ± 0.70 & 55.47 ± 0.66 & 52.95 ± 3.82 & 53.90 ± 3.04 & 62.39 ± 2.13 & 57.68 ± 0.91 & 56.73 ± 1.33 \\
KLM~\cite{mls} & 84.77 ± 2.95 & 71.07 ± 0.59 & 77.92 ± 1.31 & 73.09 ± 6.67 & 50.30 ± 7.04 & 81.80 ± 5.80 & 81.40 ± 1.58 & 71.65 ± 2.01 \\
VIM~\cite{vim} & 70.59 ± 0.43 & 54.66 ± 0.42 & 62.63 ± 0.27 & 48.32 ± 1.07 & 46.22 ± 5.46 & \textbf{46.86 ± 2.29} & 61.57 ± 0.77 & \textbf{50.74 ± 1.00} \\
KNN~\cite{knn} & 72.80 ± 0.44 & 49.65 ± 0.37 & 61.22 ± 0.14 & 48.58 ± 4.67 & 51.75 ± 3.12 & \textbf{53.56 ± 2.32} & 60.70 ± 1.03 & 53.65 ± 0.28 \\
DICE~\cite{dice} & 60.98 ± 1.10 & 54.93 ± 0.53 & 57.95 ± 0.53 & 51.79 ± 3.67 & 49.58 ± 3.32 & 64.23 ± 1.65 & 59.39 ± 1.25 & 56.25 ± 0.60 \\
RankFeat~\cite{rankfeat} & 82.78 ± 1.56 & 78.40 ± 0.95 & 80.59 ± 1.10 & 75.01 ± 5.83 & 58.49 ± 2.30 & 66.87 ± 3.80 & 77.42 ± 1.96 & 69.45 ± 1.01 \\
ASH~\cite{ash} & 68.06 ± 0.44 & 63.35 ± 0.90 & 65.71 ± 0.24 & 66.58 ± 3.88 & 46.00 ± 2.67 & 61.27 ± 2.74 & 62.95 ± 0.99 & 59.20 ± 2.46 \\
SHE~\cite{she} & 60.41 ± 0.51 & 57.74 ± 0.73 & 59.07 ± 0.25 & 58.78 ± 2.70 & 59.15 ± 7.61 & 73.29 ± 3.22 & 65.24 ± 0.98 & 64.12 ± 2.70 \\
GEN~\cite{gen} & \textbf{58.87 ± 0.69} & 49.98 ± 0.05 & \textbf{54.42 ± 0.33} & 53.92 ± 5.71 & 55.45 ± 2.76 & 61.23 ± 1.40 & 56.25 ± 1.01 & 56.71 ± 1.59 \\
ExCeL~\cite{excel} & 61.07 ± 0.81 & \textbf{49.35 ± 0.31} & 55.21 ± 0.56 & 54.67 ± 5.86 & \textbf{45.13 ± 0.33} & \textbf{51.14 ± 0.14} & 58.02 ± 1.28 & \textbf{52.24 ± 1.90} \\
\hline
\multicolumn{9}{c}{Training methods without outliers} \\
\hline
\rowcolor{gray!30}
RankOOD (Ours) & \textbf{55.71 ± 1.92} & 49.47 ± 0.80 & \textbf{52.59 ± 0.75} & \textbf{42.25 ± 2.92} & \textbf{39.21 ± 1.59} & \textbf{52.88 ± 1.89} & 55.40 ± 0.47 & \textbf{47.44 ± 0.80} \\
\hline
CRAFT~\cite{karunanayake2025craft} & 59.19 ± 0.64 & \textbf{48.26 ± 1.21} & 53.73 ± 0.62 & 48.95 ± 1.90 & 47.50 ± 5.22 & 56.97 ± 1.77 & \textbf{54.02 ± 0.30} & \textbf{51.86 ± 0.49} \\
ConfBranch~\cite{confbranch} & 74.56 ± 1.22 & 65.86 ± 0.56 & 70.21 ± 0.83 & 55.95 ± 6.15 & 76.01 ± 12.3 & 85.43 ± 1.17 & 69.90 ± 0.28 & 71.82 ± 3.39 \\
G-ODIN~\cite{godin} & 78.82 ± 1.86 & 56.34 ± 0.45 & 67.58 ± 0.98 & \textbf{27.19 ± 6.24} & \textbf{42.68 ± 5.74} & \textbf{35.83 ± 1.15} & 65.03 ± 1.16 & \textbf{42.68 ± 3.19} \\
CSI~\cite{csi} & 72.62 ± 0.49 & 67.90 ± 0.64 & 70.26 ± 0.56 & 80.54 ± 4.87 & 67.21 ± 3.35 & 90.51 ± 1.47 & 69.41 ± 0.58 & 76.92 ± 1.29 \\
ARPL~\cite{arpl} & 64.84 ± 1.25 & 58.27 ± 2.40 & 61.56 ± 1.81 & 59.12 ± 8.04 & 59.76 ± 1.58 & 71.66 ± 1.81 & 62.01 ± 0.89 & 63.14 ± 2.53 \\
MOS~\cite{mos} & 60.60 ± 1.47 & 51.49 ± 0.69 & 56.05 ± 1.01 & 52.70 ± 3.81 & 56.33 ± 8.46 & 61.24 ± 2.06 & 58.86 ± 0.41 & 57.28 ± 3.29 \\
LogitNorm~\cite{logitnorm} & 73.88 ± 1.21 & 51.89 ± 0.10 & 62.89 ± 0.57 & \textbf{34.12 ± 8.32} & 47.52 ± 8.02 & 77.38 ± 2.99 & 55.44 ± 1.45 & 53.61 ± 3.45 \\
CIDER~\cite{cider} & 82.71 ± 1.25 & 61.33 ± 0.64 & 72.02 ± 0.31 & 75.32 ± 4.21 & \textbf{17.82 ± 2.80} & 54.43 ± 2.56 & 69.30 ± 1.81 & 54.22 ± 1.24 \\
\hline
\multicolumn{9}{c}{Training methods with outliers} \\
\hline
OE~\cite{oe} & 61.26 ± 0.22 & \textbf{0.21 ± 0.01} & \textbf{30.73 ± 0.11} & 53.31 ± 9.91 & 51.84 ± 3.45 & 55.83 ± 1.82 & 58.30 ± 0.72 & 54.82 ± 2.79 \\
MCD~\cite{mcd} & 62.65 ± 0.54 & \textbf{49.10 ± 1.29} & 55.88 ± 0.85 & 62.78 ± 2.91 & \textbf{43.71 ± 3.73} & 56.89 ± 0.64 & \textbf{54.17 ± 1.13} & 54.39 ± 1.34 \\
UDG~\cite{udg} & 66.40 ± 0.51 & 56.43 ± 0.68 & 61.42 ± 0.48 & \textbf{45.14 ± 12.8} & 59.67 ± 5.62 & 71.33 ± 3.59 & 59.85 ± 0.57 & 59.00 ± 3.35 \\
MixOE~\cite{mixoe} & 61.12 ± 1.08 & \textbf{49.32 ± 0.36} & 55.22 ± 0.49 & 59.49 ± 7.74 & 73.09 ± 4.00 & 66.04 ± 0.98 & 56.93 ± 0.78 & 63.88 ± 2.48 \\
\hline
\end{tabular}
\end{table*}

\begin{table*}[h]
\centering
\scriptsize
\captionsetup{width=0.87\linewidth}
\caption{ AUROC (\% $\uparrow$) of various methods for different OOD datasets when CIFAR-100 is ID. For each column, the top five methods are marked in \textbf{bold}. \textit{RankOOD ranks within the top four methods in three out of five far-OOD detection scenarios.}}
\label{tab:auroc_cifar100}
\begin{tabular}{l|ccc|ccccc}
\hline
Method & \multicolumn{3}{c|}{Near OOD} & \multicolumn{5}{c}{Far OOD} \\
 & CIFAR-10 & TIN & Average & MNIST & SVHN & Textures & Places365 & Average\\ 
 \hline
 \multicolumn{9}{c}{Post-hoc inference methods} \\ 
\hline
OpenMax~\cite{openmax} & 74.38 ± 0.37 & 78.44 ± 0.14 & 76.41 ± 0.25 & 76.01 ± 1.39 & 82.07 ± 1.53 & 80.56 ± 0.09 & 79.29 ± 0.40 & 79.48 ± 0.41 \\
MSP~\cite{msp} & 78.47 ± 0.07 & 82.07 ± 0.17 & 80.27 ± 0.11 & 76.08 ± 1.86 & 78.42 ± 0.89 & 77.32 ± 0.71 & 79.22 ± 0.29 & 77.76 ± 0.44 \\
TempScale~\cite{tempscale} & \textbf{79.02 ± 0.06} & 82.79 ± 0.09 & 80.90 ± 0.07 & 77.27 ± 1.85 & 79.79 ± 1.05 & 78.11 ± 0.72 & 79.80 ± 0.25 & 78.74 ± 0.51 \\
ODIN~\cite{odin} & 78.18 ± 0.14 & 81.63 ± 0.08 & 79.90 ± 0.11 & 83.79 ± 1.31 & 74.54 ± 0.76 & 79.33 ± 1.08 & 79.45 ± 0.26 & 79.28 ± 0.21 \\
MDS~\cite{mds} & 55.87 ± 0.22 & 61.50 ± 0.28 & 58.69 ± 0.09 & 67.47 ± 0.81 & 70.68 ± 6.40 & 76.26 ± 0.69 & 63.15 ± 0.49 & 69.39 ± 1.39 \\
MDSEns~\cite{mds} & 43.85 ± 0.31 & 48.78 ± 0.19 & 46.31 ± 0.24 & \textbf{98.21 ± 0.78} & 53.76 ± 1.63 & 69.75 ± 1.14 & 42.27 ± 0.73 & 66.00 ± 0.69 \\
RMDS~\cite{rmds} & 77.75 ± 0.19 & 82.55 ± 0.02 & 80.15 ± 0.11 & 79.74 ± 2.49 & 84.89 ± 1.10 & \textbf{83.65 ± 0.51} & \textbf{83.40 ± 0.46} & \textbf{82.92 ± 0.42} \\
Gram~\cite{gram} & 49.41 ± 0.58 & 53.91 ± 1.58 & 51.66 ± 0.77 & 80.71 ± 4.15 & \textbf{95.55 ± 0.60} & 70.79 ± 1.32 & 46.38 ± 1.21 & 73.36 ± 1.08 \\
EBO~\cite{ebo} & \textbf{79.05 ± 0.11} & 82.76 ± 0.08 & \textbf{80.91 ± 0.08} & 79.18 ± 1.37 & 82.03 ± 1.74 & 78.35 ± 0.83 & 79.52 ± 0.23 & 79.77 ± 0.61 \\
OpenGAN~\cite{opengan} & 63.23 ± 2.44 & 68.74 ± 2.29 & 65.98 ± 1.26 & 68.14 ± 18.8 & 68.40 ± 2.15 & 65.84 ± 3.43 & 69.13 ± 7.08 & 67.88 ± 7.16 \\
GradNorm~\cite{gradnorm} & 70.32 ± 0.20 & 69.95 ± 0.79 & 70.13 ± 0.47 & 65.35 ± 1.12 & 76.95 ± 4.73 & 64.58 ± 0.13 & 69.69 ± 0.17 & 69.14 ± 1.05 \\
ReAct~\cite{react} & 78.65 ± 0.05 & 82.88 ± 0.08 & 80.77 ± 0.05 & 78.37 ± 1.59 & 83.01 ± 0.97 & 80.15 ± 0.46 & \textbf{80.03 ± 0.11} & 80.39 ± 0.49 \\
MLS~\cite{mls} & \textbf{79.21 ± 0.10} & 82.90 ± 0.05 & \textbf{81.05 ± 0.07} & 78.91 ± 1.47 & 81.65 ± 1.49 & 78.39 ± 0.84 & 79.75 ± 0.24 & 79.67 ± 0.57 \\
KLM~\cite{mls} & 73.91 ± 0.25 & 79.22 ± 0.28 & 76.56 ± 0.25 & 74.15 ± 2.59 & 79.34 ± 0.44 & 75.77 ± 0.45 & 75.70 ± 0.24 & 76.24 ± 0.52 \\
VIM~\cite{vim} & 72.21 ± 0.41 & 77.76 ± 0.16 & 74.98 ± 0.13 & 81.89 ± 1.02 & 83.14 ± 3.71 & \textbf{85.91 ± 0.78} & 75.85 ± 0.37 & 81.70 ± 0.62 \\
KNN~\cite{knn} & 77.02 ± 0.25 & \textbf{83.34 ± 0.16} & 80.18 ± 0.15 & 82.36 ± 1.52 & 84.15 ± 1.09 & \textbf{83.66 ± 0.83} & 79.43 ± 0.47 & \textbf{82.40 ± 0.17} \\
DICE~\cite{dice} & 78.04 ± 0.32 & 80.72 ± 0.30 & 79.38 ± 0.23 & 79.86 ± 1.89 & 84.22 ± 2.00 & 77.63 ± 0.34 & 78.33 ± 0.66 & 80.01 ± 0.18 \\
RankFeat~\cite{rankfeat} & 58.04 ± 2.36 & 65.72 ± 0.22 & 61.88 ± 1.28 & 63.03 ± 3.86 & 72.14 ± 1.39 & 69.40 ± 3.08 & 63.82 ± 1.83 & 67.10 ± 1.42 \\
ASH~\cite{ash} & 76.48 ± 0.30 & 79.92 ± 0.20 & 78.20 ± 0.15 & 77.23 ± 0.46 & \textbf{85.60 ± 1.40} & 80.72 ± 0.70 & 78.76 ± 0.16 & 80.58 ± 0.66 \\
SHE~\cite{she} & 78.15 ± 0.03 & 79.74 ± 0.36 & 78.95 ± 0.18 & 76.76 ± 1.07 & 80.97 ± 3.98 & 73.64 ± 1.28 & 76.30 ± 0.51 & 76.92 ± 1.16 \\
GEN~\cite{gen} & \textbf{79.38 ± 0.04} & \textbf{83.25 ± 0.13} & \textbf{81.31 ± 0.08} & 78.29 ± 2.05 & 81.41 ± 1.50 & 78.74 ± 0.81 & \textbf{80.28 ± 0.27} & 79.68 ± 0.75 \\
ExCeL~\cite{excel} & 78.14 ± 0.09 & \textbf{83.26 ± 0.03} & 80.70 ± 0.06 & 78.99 ± 1.73 & \textbf{85.91 ± 0.73} & \textbf{83.28 ± 0.58} & 79.98 ± 0.57 & \textbf{82.04 ± 0.90} \\
\hline
\multicolumn{9}{c}{Training methods without outliers} \\
\hline
\rowcolor{gray!30}
RankOOD (Ours) & 78.84 ± 0.79 & 82.50 ± 0.39 & 80.67 ± 0.40 & \textbf{84.00 ± 2.15} & \textbf{87.75 ± 1.25} & 82.04 ± 0.70 & 80.73 ± 0.46 & \textbf{83.63 ± 1.06} \\
\hline
CRAFT~\cite{karunanayake2025craft} & \textbf{78.67 ± 0.21} & 83.14 ± 0.73 & 80.90 ± 0.33 & 80.34 ± 1.84 & 85.16 ± 1.15 & 80.91 ± 0.45 & \textbf{81.71 ± 0.12} & 82.03 ± 0.34 \\
ConfBranch~\cite{confbranch} & 68.80 ± 0.73 & 74.41 ± 0.54 & 71.60 ± 0.62 & 74.29 ± 4.44 & 65.51 ± 8.07 & 65.39 ± 0.16 & 70.42 ± 0.26 & 68.90 ± 1.83 \\
G-ODIN~\cite{godin} & 73.04 ± 0.39 & 81.26 ± 0.29 & 77.15 ± 0.28 & \textbf{91.15 ± 2.86} & 83.74 ± 3.10 & \textbf{89.62 ± 0.36} & 78.17 ± 0.62 & \textbf{85.67 ± 1.58} \\
CSI~\cite{csi} & 69.50 ± 0.43 & 73.40 ± 0.13 & 71.45 ± 0.27 & 51.79 ± 6.77 & 80.24 ± 1.80 & 62.22 ± 0.98 & 70.99 ± 0.54 & 66.31 ± 1.21 \\
ARPL~\cite{arpl} & 73.38 ± 0.78 & 76.50 ± 1.11 & 74.94 ± 0.93 & 73.77 ± 5.89 & 76.45 ± 1.00 & 69.93 ± 1.33 & 74.62 ± 0.57 & 73.69 ± 1.80 \\
MOS~\cite{mos} & 78.54 ± 0.13 & 82.26 ± 0.25 & 80.40 ± 0.18 & 80.68 ± 1.65 & 81.59 ± 3.81 & 79.92 ± 0.57 & 78.50 ± 0.34 & 80.17 ± 1.21 \\
LogitNorm~\cite{logitnorm} & 74.57 ± 0.39 & 82.37 ± 0.24 & 78.47 ± 0.31 & \textbf{90.69 ± 1.38} & 82.80 ± 4.57 & 72.37 ± 0.67 & \textbf{80.25 ± 0.61} & 81.53 ± 1.26 \\
CIDER~\cite{cider} & 67.55 ± 0.60 & 78.65 ± 0.35 & 73.10 ± 0.39 & 68.14 ± 3.98 & \textbf{97.17 ± 0.34} & 82.21 ± 1.93 & 74.43 ± 0.64 & 80.49 ± 0.68 \\
\hline
\multicolumn{9}{c}{Training methods with outliers} \\
\hline
OE~\cite{oe} & 76.70 ± 0.19 & \textbf{99.89 ± 0.02} & \textbf{88.30 ± 0.10} & 80.68 ± 5.82 & 84.37 ± 1.34 & 82.18 ± 0.68 & 78.39 ± 0.41 & 81.41 ± 1.49 \\
MCD~\cite{mcd} & 75.40 ± 0.46 & 78.75 ± 0.21 & 77.07 ± 0.32 & 68.25 ± 1.99 & 75.92 ± 0.37 & 77.07 ± 0.76 & 77.65 ± 0.09 & 74.72 ± 0.78 \\
UDG~\cite{udg} & 75.15 ± 0.15 & 80.90 ± 0.21 & 78.02 ± 0.10 & \textbf{83.88 ± 5.98} & 79.80 ± 1.61 & 75.57 ± 0.80 & 79.11 ± 0.17 & 79.59 ± 1.77 \\
MixOE~\cite{mixoe} & 78.17 ± 0.29 & \textbf{83.73 ± 0.12} & \textbf{80.95 ± 0.20} & 76.06 ± 5.52 & 72.28 ± 0.81 & 77.34 ± 0.91 & 79.92 ± 0.30 & 76.40 ± 1.44 \\
\hline
\end{tabular}
\end{table*}


\begin{table*}[h]
\centering
\scriptsize
\captionsetup{width=0.87\linewidth}
\caption{FPR95 (\% $\downarrow$) of various methods for different OOD datasets when CIFAR-10 is ID. For each column, the top five methods are marked in \textbf{bold}. \textit{RankOOD ranks within the top five methods in two out of three near-OOD detection scenarios.}}
\label{tab:fpr_cifar10}
\begin{tabular}{l|ccc|ccccc}
\hline
Method & \multicolumn{3}{c|}{Near OOD} & \multicolumn{5}{c}{Far OOD} \\
 & CIFAR-100 & TIN & Average & MNIST & SVHN & Textures & Places365 & Average\\ 
 \hline
\multicolumn{9}{c}{Post-hoc inference methods} \\ 
\hline
OpenMax~\cite{openmax} & 48.06 ± 3.25 & 39.18 ± 1.44 & 43.62 ± 2.27 & 23.33 ± 4.67 & 25.40 ± 1.47 & 31.50 ± 4.05 & 38.52 ± 2.27 & 29.69 ± 1.21 \\ 
MSP~\cite{msp} & 53.08 ± 4.86 & 43.27 ± 3.00 & 48.17 ± 3.92 & 23.64 ± 5.81 & 25.82 ± 1.64 & 34.96 ± 4.64 & 42.47 ± 3.81 & 31.72 ± 1.84 \\ 
TempScale~\cite{tempscale} & 55.81 ± 5.07 & 46.11 ± 3.63 & 50.96 ± 4.32 & 23.53 ± 7.05 & 26.97 ± 2.65 & 38.16 ± 5.89 & 45.27 ± 4.50 & 33.48 ± 2.39 \\ 
ODIN~\cite{odin} & 77.00 ± 5.74 & 75.38 ± 6.42 & 76.19 ± 6.08 & 23.83 ± 12.3 & 68.61 ± 0.52 & 67.70 ± 11.1 & 70.36 ± 6.96 & 57.62 ± 4.24 \\ 
MDS~\cite{mds} & 52.81 ± 3.62 & 46.99 ± 4.36 & 49.90 ± 3.98 & 27.30 ± 3.55 & 25.96 ± 2.52 & 27.94 ± 4.20 & 47.67 ± 4.54 & 32.22 ± 3.40 \\ 
MDSEns~\cite{mds} & 91.87 ± 0.10 & 92.66 ± 0.42 & 92.26 ± 0.20 & \textbf{1.30 ± 0.51} & 74.34 ± 1.04 & 76.07 ± 0.17 & 94.16 ± 0.33 & 61.47 ± 0.48 \\ 
RMDS~\cite{rmds} & 43.86 ± 3.49 & 33.91 ± 1.39 & 38.89 ± 2.39 & 21.49 ± 2.32 & 23.46 ± 1.48 & 25.25 ± 0.53 & 31.20 ± 0.28 & 25.35 ± 0.73 \\ 
Gram~\cite{gram} & 91.68 ± 2.24 & 90.06 ± 1.59 & 90.87 ± 1.91 & 70.30 ± 8.96 & 33.91 ± 17.4 & 94.64 ± 2.71 & 90.49 ± 1.93 & 72.34 ± 6.73 \\ 
EBO~\cite{ebo} & 66.60 ± 4.46 & 56.08 ± 4.83 & 61.34 ± 4.63 & 24.99 ± 12.9 & 35.12 ± 6.11 & 51.82 ± 6.11 & 54.85 ± 6.52 & 41.69 ± 5.32 \\ 
OpenGAN~\cite{opengan} & 94.84 ± 3.83 & 94.11 ± 4.21 & 94.48 ± 4.01 & 79.54 ± 19.7 & 75.27 ± 26.9 & 83.95 ± 14.9 & 95.32 ± 4.45 & 83.52 ± 11.6 \\ 
GradNorm~\cite{gradnorm} & 94.54 ± 1.11 & 94.89 ± 0.60 & 94.72 ± 0.82 & 85.41 ± 4.85 & 91.65 ± 2.42 & 98.09 ± 0.49 & 92.46 ± 2.28 & 91.90 ± 2.23 \\ 
ReAct~\cite{react} & 67.40 ± 7.34 & 59.71 ± 7.31 & 63.56 ± 7.33 & 33.77 ± 18.0 & 50.23 ± 15.9 & 51.42 ± 11.4 & 44.20 ± 3.35 & 44.90 ± 8.37 \\ 
MLS~\cite{mls} & 66.59 ± 4.44 & 56.06 ± 4.82 & 61.32 ± 4.62 & 25.06 ± 12.9 & 35.09 ± 6.09 & 51.73 ± 6.13 & 54.84 ± 6.51 & 41.68 ± 5.27 \\ 
KLM~\cite{mls} & 90.55 ± 5.83 & 85.18 ± 7.60 & 87.86 ± 6.37 & 76.22 ± 12.1 & 59.47 ± 7.06 & 81.95 ± 9.95 & 95.58 ± 2.12 & 78.31 ± 4.84 \\ 
VIM~\cite{vim} & 49.19 ± 3.15 & 40.49 ± 1.55 & 44.84 ± 2.31 & 18.36 ± 1.42 & 19.29 ± 0.41 & \textbf{21.14 ± 1.83} & 41.43 ± 2.17 & 25.05 ± 0.52 \\ 
KNN~\cite{knn} & 37.64 ± 0.31 & 30.37 ± 0.65 & 34.01 ± 0.38 & 20.05 ± 1.36 & 22.60 ± 1.26 & 24.06 ± 0.55 & 30.38 ± 0.63 & 24.27 ± 0.40 \\ 
DICE~\cite{dice} & 73.71 ± 7.67 & 66.37 ± 7.68 & 70.04 ± 7.64 & 30.83 ± 10.5 & 36.61 ± 4.74 & 62.42 ± 4.79 & 77.19 ± 12.6 & 51.76 ± 4.42 \\ 
RankFeat~\cite{rankfeat} & 65.32 ± 3.48 & 56.44 ± 5.76 & 60.88 ± 4.60 & 61.86 ± 12.8 & 64.49 ± 7.38 & 59.71 ± 9.79 & 43.70 ± 7.39 & 57.44 ± 7.99 \\ 
ASH~\cite{ash} & 87.31 ± 2.06 & 86.25 ± 1.58 & 86.78 ± 1.82 & 70.00 ± 10.6 & 83.64 ± 6.48 & 84.59 ± 1.74 & 77.89 ± 7.28 & 79.03 ± 4.22 \\ 
SHE~\cite{she} & 81.00 ± 3.42 & 78.30 ± 3.52 & 79.65 ± 3.47 & 42.22 ± 20.6 & 62.74 ± 4.01 & 84.60 ± 5.30 & 76.36 ± 5.32 & 66.48 ± 5.98 \\ 
GEN~\cite{gen} & 58.75 ± 3.97 & 48.59 ± 2.34 & 53.67 ± 3.14 & 23.00 ± 7.75 & 28.14 ± 2.59 & 40.74 ± 6.61 & 47.03 ± 3.22 & 34.73 ± 1.58 \\ 
ExCeL~\cite{excel} & 71.16 ± 1.34 & 61.42 ± 0.26 & 66.55 ± 0.43 & \textbf{15.46 ± 1.89} & 31.78 ± 3.65 & 53.67 ± 2.19 & 55.09 ± 1.12 & 40.03 ± 0.84 \\ 
\hline
\multicolumn{9}{c}{Training methods without outliers} \\
\hline
\rowcolor{gray!30}
RankOOD (Ours) & \textbf{35.42 ± 0.65} & 28.01 ± 1.41 & \textbf{31.72 ± 0.67} & 17.18 ± 4.09 & 15.75 ± 3.20 & 22.66 ± 5.39 & 28.27 ± 0.83 & 20.96 ± 2.55 \\ 
\hline
CRAFT~\cite{karunanayake2025craft} & 36.61 ± 2.93 & \textbf{27.28 ± 0.09} & 31.94 ± 1.41 & 17.13 ± 0.99 & 14.58 ± 4.62 & \textbf{20.78 ± 0.04} & \textbf{25.12 ± 0.15} & \textbf{19.40 ± 0.88} \\ 
ConfBranch~\cite{confbranch} & \textbf{34.44 ± 0.81} & \textbf{28.11 ± 0.61} & \textbf{31.28 ± 0.66} & \textbf{15.79 ± 2.00} & \textbf{14.06 ± 0.84} & 27.24 ± 1.32 & 28.85 ± 1.03 & 21.48 ± 0.94 \\ 
G-ODIN~\cite{godin} & 48.86 ± 2.91 & 42.21 ± 2.18 & 45.54 ± 2.52 & \textbf{4.53 ± 2.08} & \textbf{10.72 ± 0.88} & 27.27 ± 6.73 & 43.30 ± 3.57 & 21.45 ± 1.91 \\ 
CSI~\cite{csi} & 37.57 ± 0.89 & 29.74 ± 0.42 & 33.66 ± 0.64 & 24.41 ± 1.57 & 17.56 ± 0.12 & 28.95 ± 1.33 & 34.76 ± 1.52 & 26.42 ± 0.29 \\ 
ARPL~\cite{arpl} & 43.38 ± 0.37 & 37.28 ± 1.21 & 40.33 ± 0.70 & 21.49 ± 2.03 & 35.68 ± 3.48 & 35.19 ± 1.79 & 37.21 ± 0.80 & 32.39 ± 0.74 \\ 
MOS~\cite{mos} & 79.38 ± 5.06 & 78.05 ± 6.69 & 78.72 ± 5.86 & 65.95 ± 17.5 & 57.79 ± 5.79 & 76.78 ± 3.86 & 51.09 ± 1.33 & 62.90 ± 6.62 \\ 
LogitNorm~\cite{logitnorm} & \textbf{34.37 ± 1.30} & \textbf{24.30 ± 0.54} & \textbf{29.34 ± 0.81} & \textbf{3.93 ± 1.99} & \textbf{8.33 ± 1.78} & \textbf{21.94 ± 0.85} & \textbf{21.04 ± 0.71} & \textbf{13.81 ± 0.20} \\ 
CIDER~\cite{cider} & \textbf{35.60 ± 0.78} & 28.61 ± 1.10 & 32.11 ± 0.94 & 24.76 ± 2.82 & \textbf{8.04 ± 0.43} & 25.05 ± 3.29 & \textbf{25.03 ± 1.36} & \textbf{20.72 ± 0.85} \\ 
\hline
\multicolumn{9}{c}{Training methods with outliers} \\
\hline
OE~\cite{oe} & 36.71 ± 2.06 & \textbf{2.97 ± 1.17} & \textbf{19.84 ± 0.95} & 24.67 ± 2.55 & \textbf{1.25 ± 0.36} & \textbf{12.07 ± 2.14} & \textbf{14.53 ± 2.80} & \textbf{13.13 ± 0.53} \\ 
MCD~\cite{mcd} & \textbf{34.36 ± 0.37} & \textbf{25.98 ± 0.44} & \textbf{30.17 ± 0.06} & 62.11 ± 11.8 & 19.43 ± 5.93 & 22.51 ± 5.16 & \textbf{24.10 ± 1.58} & 32.03 ± 4.21 \\ 
UDG~\cite{udg} & 40.75 ± 0.69 & 29.93 ± 1.27 & 35.34 ± 0.95 & 16.61 ± 5.14 & 17.39 ± 7.87 & \textbf{19.70 ± 1.89} & 27.70 ± 1.80 & \textbf{20.35 ± 2.41} \\ 
MixOE~\cite{mixoe} & 58.29 ± 8.25 & 44.62 ± 7.57 & 51.45 ± 7.78 & 38.28 ± 13.4 & 20.36 ± 3.99 & 33.19 ± 4.28 & 43.54 ± 4.95 & 33.84 ± 4.77 \\ 
\hline
\end{tabular}
\end{table*}

\begin{table*}[h]
\centering
\scriptsize
\captionsetup{width=0.87\linewidth}
\caption{AUROC (\% $\uparrow$) of various methods for different OOD datasets when CIFAR-10 is ID. For each column, the top five methods are marked in \textbf{bold}. \textit{RankOOD achieves on par performance compared to CRAFT, a class rank based method.}}
\label{tab:auroc_cifar10}
\begin{tabular}{l|ccc|ccccc}
\hline
Method & \multicolumn{3}{c|}{Near OOD} & \multicolumn{5}{c}{Far OOD} \\
 & CIFAR-100 & TIN & Average & MNIST & SVHN & Textures & Places365 & Average\\ 
 \hline
\multicolumn{9}{c}{Post-hoc inference methods} \\ 
\hline
OpenMax~\cite{openmax} & 86.91 ± 0.31 & 88.32 ± 0.28 & 87.62 ± 0.29 & 90.50 ± 0.44 & 89.77 ± 0.45 & 89.58 ± 0.60 & 88.63 ± 0.28 & 89.62 ± 0.19 \\
MSP~\cite{msp} & 87.19 ± 0.33 & 88.87 ± 0.19 & 88.03 ± 0.25 & 92.63 ± 1.57 & 91.46 ± 0.40 & 89.89 ± 0.71 & 88.92 ± 0.47 & 90.73 ± 0.43 \\
TempScale~\cite{tempscale} & 87.17 ± 0.40 & 89.00 ± 0.23 & 88.09 ± 0.31 & 93.11 ± 1.77 & 91.66 ± 0.52 & 90.01 ± 0.74 & 89.11 ± 0.52 & 90.97 ± 0.52 \\
ODIN~\cite{odin} & 82.18 ± 1.87 & 83.55 ± 1.84 & 82.87 ± 1.85 & 95.24 ± 1.96 & 84.58 ± 0.77 & 86.94 ± 2.26 & 85.07 ± 1.24 & 87.96 ± 0.61 \\
MDS~\cite{mds} & 83.59 ± 2.27 & 84.81 ± 2.53 & 84.20 ± 2.40 & 90.10 ± 2.41 & 91.18 ± 0.47 & 92.69 ± 1.06 & 84.90 ± 2.54 & 89.72 ± 1.36 \\
MDSEns~\cite{mds} & 61.29 ± 0.23 & 59.57 ± 0.53 & 60.43 ± 0.26 & \textbf{99.17 ± 0.41} & 66.56 ± 0.58 & 77.40 ± 0.28 & 52.47 ± 0.15 & 73.90 ± 0.27 \\
RMDS~\cite{rmds} & 88.83 ± 0.35 & 90.76 ± 0.27 & 89.80 ± 0.28 & 93.22 ± 0.80 & 91.84 ± 0.26 & 92.23 ± 0.23 & 91.51 ± 0.11 & 92.20 ± 0.21 \\
Gram~\cite{gram} & 58.33 ± 4.49 & 58.98 ± 5.19 & 58.66 ± 4.83 & 72.64 ± 2.34 & 91.52 ± 4.45 & 62.34 ± 8.27 & 60.44 ± 3.41 & 71.73 ± 3.20 \\
EBO~\cite{ebo} & 86.36 ± 0.58 & 88.80 ± 0.36 & 87.58 ± 0.46 & 94.32 ± 2.53 & 91.79 ± 0.98 & 89.47 ± 0.70 & 89.25 ± 0.78 & 91.21 ± 0.92 \\
OpenGAN~\cite{opengan} & 52.81 ± 7.69 & 54.62 ± 7.68 & 53.71 ± 7.68 & 56.14 ± 24.1 & 52.81 ± 27.6 & 56.14 ± 18.3 & 53.34 ± 5.79 & 54.61 ± 15.5 \\
GradNorm~\cite{gradnorm} & 54.43 ± 1.59 & 55.37 ± 0.41 & 54.90 ± 0.98 & 63.72 ± 7.37 & 53.91 ± 6.36 & 52.07 ± 4.09 & 60.50 ± 5.33 & 57.55 ± 3.22 \\
ReAct~\cite{react} & 85.93 ± 0.83 & 88.29 ± 0.44 & 87.11 ± 0.61 & 92.81 ± 3.03 & 89.12 ± 3.19 & 89.38 ± 1.49 & 90.35 ± 0.78 & 90.42 ± 1.41 \\
MLS~\cite{mls} & 86.31 ± 0.59 & 88.72 ± 0.36 & 87.52 ± 0.47 & 94.15 ± 2.48 & 91.69 ± 0.94 & 89.41 ± 0.71 & 89.14 ± 0.76 & 91.10 ± 0.89 \\
KLM~\cite{mls} & 77.89 ± 0.75 & 80.49 ± 0.85 & 79.19 ± 0.80 & 85.00 ± 2.04 & 84.99 ± 1.18 & 82.35 ± 0.33 & 78.37 ± 0.33 & 82.68 ± 0.21 \\
VIM~\cite{vim} & 87.75 ± 0.28 & 89.62 ± 0.33 & 88.68 ± 0.28 & 94.76 ± 0.38 & 94.50 ± 0.48 & \textbf{95.15 ± 0.34} & 89.49 ± 0.39 & 93.48 ± 0.24 \\
KNN~\cite{knn} & \textbf{89.73 ± 0.14} & 91.56 ± 0.26 & 90.64 ± 0.20 & 94.26 ± 0.38 & 92.67 ± 0.30 & 93.16 ± 0.24 & \textbf{91.77 ± 0.23} & 92.96 ± 0.14 \\
DICE~\cite{dice} & 77.01 ± 0.88 & 79.67 ± 0.87 & 78.34 ± 0.79 & 90.37 ± 5.97 & 90.02 ± 1.77 & 81.86 ± 2.35 & 74.67 ± 4.98 & 84.23 ± 1.89 \\
RankFeat~\cite{rankfeat} & 77.98 ± 2.24 & 80.94 ± 2.80 & 79.46 ± 2.52 & 75.87 ± 5.22 & 68.15 ± 7.44 & 73.46 ± 6.49 & 85.99 ± 3.04 & 75.87 ± 5.06 \\
ASH~\cite{ash} & 74.11 ± 1.55 & 76.44 ± 0.61 & 75.27 ± 1.04 & 83.16 ± 4.66 & 73.46 ± 6.41 & 77.45 ± 2.39 & 79.89 ± 3.69 & 78.49 ± 2.58 \\
SHE~\cite{she} & 80.31 ± 0.69 & 82.76 ± 0.43 & 81.54 ± 0.51 & 90.43 ± 4.76 & 86.38 ± 1.32 & 81.57 ± 1.21 & 82.89 ± 1.22 & 85.32 ± 1.43 \\
GEN~\cite{gen} & 87.21 ± 0.36 & 89.20 ± 0.25 & 88.20 ± 0.30 & 93.83 ± 2.14 & 91.97 ± 0.66 & 90.14 ± 0.76 & 89.46 ± 0.65 & 91.35 ± 0.69 \\
ExCeL~\cite{excel} & 85.31 ± 0.26 & 88.48 ± 0.19 & 86.89 ± 0.23 & \textbf{95.87 ± 0.45} & 91.40 ± 1.43 & 89.66 ± 0.64 & 89.84 ± 0.41 & 91.69 ± 0.18 \\
\hline
\multicolumn{9}{c}{Training methods without outliers} \\
\hline
\rowcolor{gray!30}
RankOOD (Ours) & 89.11 ± 0.24 & 91.32 ± 0.58 & 90.21 ± 0.41 & 93.95 ± 2.23 & 94.70 ± 1.03 & 92.63 ± 1.53 & 91.51 ± 0.36 & 93.19 ± 0.84 \\
\hline
CRAFT~\cite{karunanayake2025craft} & \textbf{90.18 ± 0.14} & \textbf{92.04 ± 0.06} & \textbf{91.11 ± 0.04} & 94.59 ± 0.02 & 94.94 ± 1.10 & 93.46 ± 0.29 & \textbf{92.77 ± 0.10} & 93.94 ± 0.20 \\
ConfBranch~\cite{confbranch} & 88.91 ± 0.25 & 90.77 ± 0.25 & 89.84 ± 0.24 & 94.49 ± 0.77 & \textbf{95.42 ± 0.35} & 91.10 ± 0.41 & 90.39 ± 0.40 & 92.85 ± 0.29 \\
G-ODIN~\cite{godin} & 88.14 ± 0.60 & 90.09 ± 0.54 & 89.12 ± 0.57 & \textbf{98.95 ± 0.53} & \textbf{97.76 ± 0.14} & \textbf{95.02 ± 1.10} & 90.31 ± 0.65 & \textbf{95.51 ± 0.31} \\
CSI~\cite{csi} & 88.16 ± 0.16 & 90.87 ± 0.23 & 89.51 ± 0.19 & 92.55 ± 1.15 & 95.18 ± 0.45 & 90.71 ± 0.44 & 89.56 ± 0.51 & 92.00 ± 0.30 \\
ARPL~\cite{arpl} & 86.76 ± 0.16 & 88.12 ± 0.14 & 87.44 ± 0.15 & 92.62 ± 0.88 & 87.69 ± 0.97 & 88.57 ± 0.43 & 88.39 ± 0.16 & 89.31 ± 0.32 \\
MOS~\cite{mos} & 70.57 ± 3.04 & 72.34 ± 3.16 & 71.45 ± 3.09 & 74.81 ± 10.1 & 73.66 ± 9.14 & 70.35 ± 3.11 & 86.81 ± 1.85 & 76.41 ± 5.93 \\
LogitNorm~\cite{logitnorm} & \textbf{90.95 ± 0.22} & \textbf{93.70 ± 0.06} & \textbf{92.33 ± 0.08} & \textbf{99.14 ± 0.45} & \textbf{98.25 ± 0.41} & \textbf{94.77 ± 0.43} & \textbf{94.79 ± 0.16} & \textbf{96.74 ± 0.06} \\
CIDER~\cite{cider} & 89.47 ± 0.19 & \textbf{91.94 ± 0.19} & \textbf{90.71 ± 0.16} & 93.30 ± 1.08 & \textbf{98.06 ± 0.07} & 93.71 ± 0.39 & \textbf{93.77 ± 0.68} & \textbf{94.71 ± 0.36} \\
\hline
\multicolumn{9}{c}{Training methods with outliers} \\
\hline
OE~\cite{oe} & \textbf{90.54 ± 0.53} & \textbf{99.11 ± 0.34} & \textbf{94.82 ± 0.21} & 90.22 ± 1.31 & \textbf{99.60 ± 0.14} & \textbf{97.58 ± 0.27} & \textbf{96.58 ± 0.70} & \textbf{96.00 ± 0.13} \\
MCD~\cite{mcd} & \textbf{89.88 ± 0.07} & \textbf{92.18 ± 0.18} & \textbf{91.03 ± 0.12} & 84.22 ± 2.10 & 93.76 ± 2.30 & 93.35 ± 1.30 & 92.66 ± 0.36 & 91.00 ± 1.10 \\
UDG~\cite{udg} & 88.62 ± 0.32 & 91.20 ± 0.20 & 89.91 ± 0.25 & \textbf{95.81 ± 1.52} & 94.55 ± 2.27 & \textbf{93.92 ± 0.44} & 91.97 ± 0.41 & \textbf{94.06 ± 0.90} \\
MixOE~\cite{mixoe} & 87.47 ± 0.97 & 90.00 ± 0.73 & 88.73 ± 0.82 & 91.66 ± 2.21 & 93.82 ± 1.27 & 91.84 ± 0.51 & 90.38 ± 0.55 & 91.93 ± 0.69 \\
\hline
\end{tabular}
\end{table*}

\subsection{Effectiveness of RankOOD-T}
\label{sec:ood_scores}

\begin{figure}[h]
    \centering
    \includegraphics[width=0.90\linewidth]{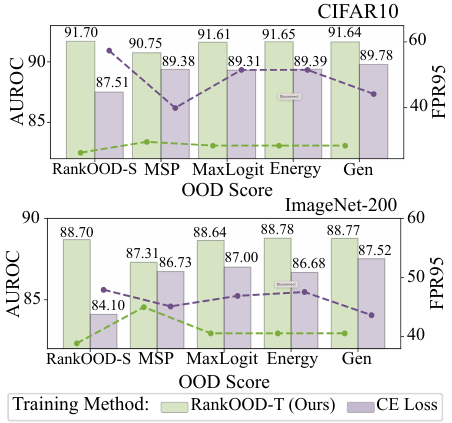}
    \caption{Avg. AUROC (bars, left y-axis) and FPR95 (lines, right y-axis) across multiple OOD datasets for other scoring functions.}
    \label{fig:diffrent_ood_scores}
    \vspace{-4mm}
\end{figure}

Fig.~\ref{fig:diffrent_ood_scores} shows standard logit-based OOD scores (MSP, MaxLogit, EBO (Energy based), and GEN) under RankOOD-T. It can be seen that RankOOD-T performs better across all scoring methods, including RankOOD-S defined in Eq.~6
, compared to cross-entropy (CE) training. \emph{This indicates that the observed performance gains primarily stem from the RankOOD-T objective rather than the RankOOD-S function itself.}  
In particular, ListMLE training improve s AUROC by at least 1.5\% and reduces FPR95 by over 7\% compared to CE training. Nonetheless, as shown in Fig~\ref{fig:diffrent_ood_scores}, logit-based OOD scores can be used in open-world settings to obtain on-par performance after RankOOD-T.

\subsection{Scalability and Computational Cost}
\label{sec:gputime_ilpruntime}
\noindent\textbf{GPU Time: }~\cref{tab:gpu_time} reports per-epoch GPU training time (seconds). As the number of classes increases, all methods exhibit comparable training costs. RankOOD incurs approximately 34\% higher GPU time than LogitNorm in TinyImageNet due to the ListMLE objective, which requires computing probabilities over full permutations. \emph{We clarify that RankOOD-T follows the same pre-training paradigm as LogitNorm with a total of 300/500 epochs and doesn't need extra fine-tuning epochs like CRAFT as canonical classes can be derived from a SOTA pre-trained model.}

\noindent\textbf{ILP Runtime: }Tab.~\ref{tab:gpu_time} reports ILP rank execution time. ILP time grows exponentially with the number of ranks and classes, with worst-case complexity $O(2^{C \times K})$ where $C$-classes and $K$-ranks. However, this can be addressed using a greedy approach with complexity $O(CK.log(K))$, significantly reducing computational cost.



\subsection{RankOOD Score Example}
\label{sec:appendix_score_example}
\begin{figure}[h]
    \centering
    \includegraphics[width=0.92\linewidth]{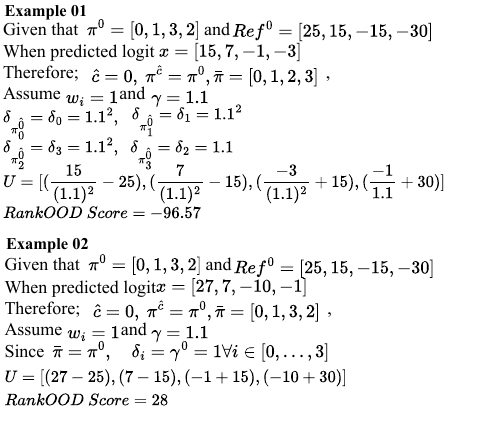}
    \caption{Toy RankOOD-S computation for a four-class problem.}
    \label{fig:ood_score_example}
    \vspace{-4mm}
\end{figure}

We provide a step-by-step examples of RankOOD-S computation in \cref{fig:ood_score_example}.
\vspace{-3mm}

\subsection{Conditional Probability Matrices}
\label{sec:appendix_cp}

\begin{figure}[t]
    \centering
    \includegraphics[width=0.97\linewidth]{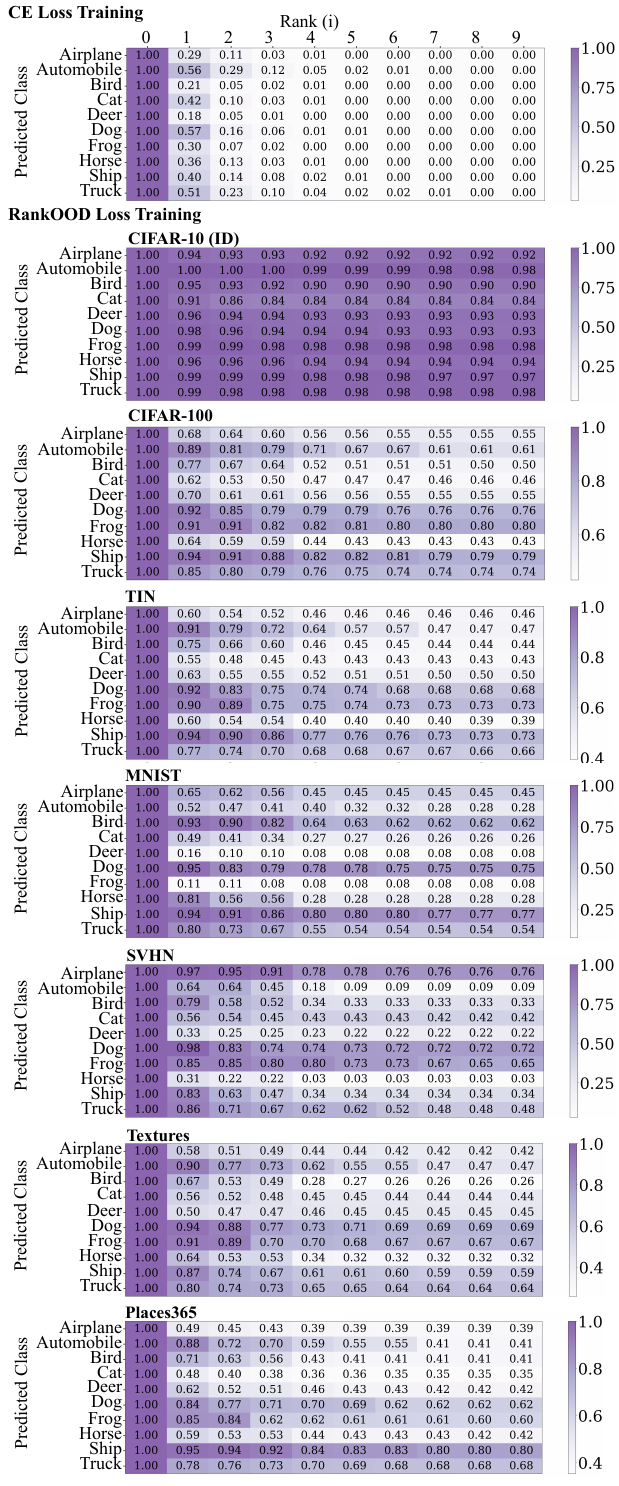}
    \caption{CIFAR-10 Conditional probability
matrix (CP) of rank position i given that all prior ranks have been correctly predicted}
    \label{fig:cpm_cifar10_all}
\end{figure}

In Sec.~5.3
, we reported class-conditional probability (CP) matrices for only five CIFAR-10 classes. Here, we provide the complete CP matrices for all ten classes under both CE and RankOOD training. For RankOOD, we additionally report CP matrices for all OOD datasets as well as the ID dataset (CIFAR-10). As shown in Fig.~\ref{fig:cpm_cifar10_all}, each matrix reflects the model’s ability to preserve the class-wise canonical rank order. 
Under conventional CE training, a sample predicted as Airplane achieves a probability of $0.29$ for correctly identifying the rank-4 label, conditioned on all preceding ranks (rank-1 through rank-3) being accurately predicted. In contrast, RankOOD training substantially improves the model’s ability to maintain deeper ranking consistency on in-distribution (ID) data: for samples classified as Airplane, the conditional probability of correctly predicting the rank-9 label increases to $0.92$.
However, when evaluated on an OOD dataset such as CIFAR-100, this probability decreases to $0.55$. This pronounced drop highlights that OOD examples are significantly less likely to preserve the learned ranking structure, thereby providing an effective signal for distinguishing ID from OOD samples.

\end{document}